\documentclass{acm_proc_article-sp}
\usepackage{amsbsy}
\usepackage{mathrsfs}
\usepackage{amssymb}
\usepackage{bm}
\usepackage{amsmath,graphicx}
\usepackage{array,booktabs}
\usepackage{verbatim}
\usepackage{multirow}
\usepackage{subfigure}
\usepackage{url}
\usepackage{graphicx}

\DeclareMathOperator*{\argmax}{\mathrm{argmax}}

\newcommand{\commentout}[1]{%
}

\newcommand{\secref}[1]{Section~\ref{#1}}
\newcommand{\appref}[1]{Appendix~\ref{#1}}
\newcommand{\figref}[1]{Figure~\ref{#1}}
\newcommand{\tabref}[1]{Table~\ref{#1}}

\newcommand{\PHITS}{\mbox{\textsc{phits}}}
\newcommand{\PLSA}{\mbox{\textsc{plsa}}}
\newcommand{\LDA}{\mbox{\textsc{lda}}}
\newcommand{\LTHM}{\mbox{\textsc{lthm}}}
\newcommand{\RTM}{\mbox{\textsc{rtm}}}
\newcommand{\MMSBM}{\mbox{\textsc{mmsb}}}
\newcommand{\BKN}{\mbox{\textsc{bkn}}}
\newcommand{\MTLM}{\mbox{\textsc{pmtlm}}}
\newcommand{\MTLMDC}{\mbox{\textsc{pmtlm-dc}}}
\newcommand{\PHITSPLSA}{\mbox{\textsc{phits-plsa}}}
\newcommand{\LINKLDA}{\mbox{\textsc{link-lda}}}
\newcommand{\LINKPLSALDA}{\mbox{\textsc{link-plsa-lda}}}
\newcommand{\PAIRLINKLDA}{\mbox{\textsc{pairwise link-lda}}}
\newcommand{\CPLDC}{\mbox{\textsc{c-pldc}}}
\newcommand{\HTM}{\mbox{\textsc{htm}}}
\newcommand{\KL}{\mbox{\textsc{kl}}}
\newcommand{\Multi}{\textrm{Multi}}
\newcommand{\Poi}{\textrm{Poi}}
\newcommand{\Bin}{\textrm{Bin}}
\newcommand{\EM}{\textrm{EM}}
\newcommand{\NMI}{\textrm{NMI}}
\newcommand{\VI}{\textrm{VI}}
\newcommand{\PWF}{\textrm{PWF}}
\newcommand{\MI}{\textrm{MI}}
\newcommand{\entropy}{\textrm{H}}
\newcommand{\Llinks}{\mathcal{L}^{\textrm{links}}}
\newcommand{\Lcontent}{\mathcal{L}^{\textrm{content}}}
\newcommand{\hattheta}{\widehat{\theta}}
\newcommand{\hatbeta}{\widehat{\beta}}
\newcommand{\hateta}{\widehat{\eta}}

\begin{document}
%

\title{Scalable Text and Link Analysis with Mixed-Topic Link Models
}
%
%
%
%
%

\numberofauthors{4} 
%
\author{
%
%
\alignauthor
Yaojia Zhu\\
       \affaddr{University of New Mexico}\\
       \email{yaojia.zhu@gmail.com}
\alignauthor
Xiaoran Yan\\
       \affaddr{University of New Mexico}\\
       \email{everyxt@gmail.com}
\alignauthor
Lise Getoor\\
       \affaddr{University of Maryland}\\
       \email{getoor@cs.umd.edu}
\and  
\alignauthor Cristopher Moore\\
       \affaddr{Santa Fe Institute}\\
       \email{moore@santafe.edu}
}

\maketitle

\begin{abstract}
Many data sets contain rich information about
objects,
as well as
pairwise relations between them.  For instance, in networks of websites,
scientific papers, and other documents,
%
each node has content consisting of a collection of words,
as well as hyperlinks or citations to other nodes.
%
In order to perform inference on such data sets,
and make \mbox{predictions} and recommendations, it
is useful to have models that
are able to capture the processes which
generate the text at each node and the links between them.
%
In this paper, we combine classic ideas in topic modeling
with a variant of the mixed-membership block model recently developed in the
statistical physics community.  The resulting
model has the advantage that its parameters, including the mixture of topics
of each document and the resulting overlapping communities,
can be inferred with a simple and scalable expectation-maximization algorithm.
We test our model on three data sets, performing unsupervised topic classification and link prediction.
For both tasks, our model outperforms several existing state-of-the-art methods, achieving higher accuracy
with significantly less computation, analyzing a data set with 1.3 million words and
44 thousand links in a few minutes.
\end{abstract}






\section{Introduction}

Many modern data sets contain not only rich information about each object, but also pairwise relationships between them, forming networks where each object is a node and links represent the relationships.
In document networks, for example, each node is a document containing a sequence of words, and the
links between nodes are citations or hyperlinks.
Both the content of the documents and the topology of the links between them are meaningful.

Over the past few years, two disparate communities have been
approaching these data sets from different points of view.  In the
data mining community,
the goal has been to augment traditional
approaches to learning and data mining by including relations between
objects~\cite{getoor2005link,yu2010link}
for instance, to use the links between documents to help us label them by topic.  In the network community, including its subset in statistical physics, the goal has been to augment traditional community structure algorithms such as the stochastic block model~\cite{FIENBERG1981,HOLLAND1983,SNIJDERS1997} by taking node attributes into account: for instance, to use the content of documents, rather than just the topological links between them, to help us understand their community structure.


In the original stochastic block model, each node has a discrete label, assigning it to one of $k$ communities.  These labels, and the $k \times k$ matrix of probabilities with which a given pair of nodes with a given pair of labels have a link between them, can be inferred using Monte Carlo algorithms~(e.g. \cite{Moore2011}) or, more efficiently, with belief propagation~\cite{LenkaBP1,LenkaBP2} or pseudolikelihood approaches~\cite{pseudolikelihood}.  However, in real networks communities often overlap, and a given node can belong to multiple communities.  This led to the \emph{mixed-membership} block model~\cite{Airoldi2008}, where the goal is to infer, for each node $v$, a distribution or mixture of labels $\theta_v$ describing to what extent it belongs to each community.  If we assume that links are assortative, i.e., that nodes are more likely to link to others in the same community, then the probability of a link between two nodes $v$ and $v'$ depends on some measure of similarity (say, the inner product) of $\theta_v$ and $\theta_{v'}$.

These mixed-membership block models fit nicely with classic ideas in topic modeling.  In models such as Probabilistic Latent Semantic Analysis (\PLSA)~\cite{Hofmann1999} and Latent Dirichlet Allocation (\LDA)~\cite{Blei2003}, each document $d$ has a mixture $\theta_d$ of topics.  Each topic corresponds in turn to a probability distribution over words, and each word in $d$ is generated independently from the resulting mixture of distributions.  If we think of $\theta_d$ as both the mixture of topics for generating words and the mixture of communities for generating links, then we can infer $\{ \theta_d \}$ jointly from the documents' content and the presence or absence of links between them.

There are many possible such models, and we are far from the first to think along these lines.  Our innovation is to take as our starting point a particular mixed-membership block model recently developed in the statistical physics community~\cite{Ball2011}, which we refer to as the \BKN\ model.  It differs from the mixed-membership stochastic block model (\MMSBM) of~\cite{Airoldi2008} in several ways:
\begin{enumerate}
\item The \BKN\ model treats the community membership mixtures $\theta_d$ directly as parameters to be inferred.  In contrast, \MMSBM\ treats $\theta_d$ as hidden variables generated by a Dirichlet distribution, and infers the hyperparameters of that distribution.  The situation between \PLSA\ and \LDA\ is similar; \PLSA\ infers the topic mixtures $\theta_d$, while \LDA\ generates them from a Dirichlet distribution.
\item The \MMSBM\ model generates each link according to a Bernoulli distribution, with an extra parameter for sparsity.  Instead, \BKN\ treats the links as a random multigraph, where the number of links $A_{dd'}$ between each pair of nodes is Poisson-distributed.  As a result, the derivatives of the log-likelihood with respect to $\theta_d$ and the other parameters are particularly simple.
\end{enumerate}
These two factors make it possible to fit the \BKN\ model using an efficient and exact expectation-maximization (\EM) algorithm, making its inference highly scalable.  The \BKN\ model has another advantage as well:
\begin{enumerate}
\item[3.] The \BKN\ model is \emph{degree-corrected}, in that it takes the observed degrees of the nodes into account when computing the expected number of edges between them.  Thus it recognizes that two documents that have very different degrees might in fact have the same mix of topics; one may simply be more popular than the other.
\end{enumerate}

In our work, we use a slight variant of the \BKN\ model to generate the links, and we use \PLSA\ to generate the text.  We present an \EM\ algorithm for inferring the topic mixtures and other parameters.  (While we do not impose a Dirichlet prior on the topic mixtures, it is easy to add a corresponding term to the update equations.)  Our algorithm is scalable in the sense that each iteration takes $O(K(N+M+R))$ time for networks with $K$ topics, $N$ documents, and $M$ links, where $R$ is the sum over documents of the number of distinct words appearing in each one.  In practice, our \EM\ algorithm converges within a small number of iterations, making the total running time linear in the size of the corpus.

Our model can be used for a variety of learning and generalization tasks, including document classification or link prediction.  For document classification, we can obtain hard labels for each document by taking its most-likely topic with respect to $\theta_d$, and optionally improve
these labels further with local search.
For link prediction, we train the model using a subset of the links, and then ask it to rank the remaining pairs of documents according to the probability of a link between them.  For each task we determine the optimal relative weight of the content vs.\ the link information.

We performed experiments on three real-world data sets, with thousands of documents and millions of words.  Our results show that our algorithm is more accurate, and considerably faster, than previous techniques for both document classification and link prediction.

The rest of the paper is organized as follows.  \secref{sec:model}~describes our generative model, and compares it
with related models in the literature. \secref{sec:algorithm}~gives our \EM\
algorithm and analyzes its running time.  \secref{sec:experiments}~contains our
experimental results for document classification and link prediction, comparing our accuracy and running time
with other techniques.  In~\secref{sec:conclusion}, we conclude, and
offer some directions for further work.


\section{Our Model and Previous Work}
\label{sec:model}

In this section, we give our proposed model, which we call the \emph{Poisson mixed-topic link model} (\MTLM) and its degree-corrected variant \MTLMDC.

\subsection{The Generative Model}

Consider a network of $N$ documents.  Each document $d$ has a fixed length $L_d$, and consists of a string of words $w_{d\ell}$ for $1 \le \ell \le L_d$, where $1 \le w_{d\ell} \le W$ where $W$ is the number of distinct words.  In addition, each pair of documents $d, d'$ has an integer number of links connecting them, giving an adjacency matrix $A_{dd'}$.  There are $K$ topics, which play the dual role of the overlapping communities in the network.

Our model generates both the content $\{ w_{d\ell} \}$ and the links $\{ A_{dd'} \}$ as follows.  We generate the content using the \PLSA\ model~\cite{Hofmann1999}.  Each topic $z$ is associated with a probability distribution $\beta_z$ over words, and each document has a probability distribution $\theta_d$ over topics.
For each document $1 \le d \le N$ and each $1 \le \ell \le L_d$, we independently
  \begin{enumerate}
   \item choose a topic $z = z_{d\ell} \sim \Multi(\theta_d)$, and
   \item choose the word $w_{d\ell} \sim \Multi(\beta_z)$.
  \end{enumerate}
Thus the total probability that $w_{d\ell}$ is a given word $w$ is
\begin{equation}
\label{eq:word_gen_prob}
\Pr[w_{d\ell} = w] = \sum_{z=1}^K \theta_{dz} \beta_{zw} \, .
\end{equation}
We assume that the number of topics $K$ is fixed.  The distributions $\beta_z$ and $\theta_d$ are parameters to be inferred.

We generate the links using a version of the Ball-Karrer-Newman (\BKN) model \cite{Ball2011}.  Each topic $z$ is associated with a link density $\eta_z$.  For each pair of documents $d, d'$ and each topic $z$, we independently generate a number of links which is Poisson-distributed with mean $\theta_{dz} \theta_{d'z} \eta_z$.  Since the sum of independent Poisson variables is Poisson, the total number of links between $d$ and $d'$ is distributed as
\begin{equation}
\label{eq:link_poission_vanilla}
A_{dd'} \sim \Poi\left( \sum_z \theta_{dz} \theta_{d'z} \eta_z \right) \, .
\end{equation}

Since $A_{dd'}$ can exceed $1$, this gives a random multigraph.  In the data sets we study below, $A_{dd'}$ is $1$ or $0$ depending on whether $d$ cites $d'$, giving a simple graph.  On the other hand, in the sparse case the event that $A_{dd'} > 1$ has low probability in our model.  Moreover, the fact that $A_{dd'}$ is Poisson-distributed rather than Bernoulli makes the derivatives of the likelihood with respect to the parameters $\theta_{dz}$ and $\eta_z$ very simple, allowing us to write down an efficient \EM\ algorithm for inferring them.


This version of the model assumes that links are assortative, i.e., that links between documents only form to the extent that they belong to the same topic.  One can easily generalize the model to include disassortative links as well, replacing $\eta_z$ with a matrix $\eta_{zz'}$ that allows documents with distinct topics $z, z'$ to link~\cite{Ball2011}.

We also consider \emph{degree-corrected} versions of this model, where in addition to its topic mixture $\theta_d$, each document has a propensity $S_d$ of forming links.  In that case,
\begin{equation}
\label{eq:link_poission_dc}
A_{dd'} \sim \Poi\left( S_d S_{d'} \sum_z \theta_{dz} \theta_{d'z} \eta_z \right) \, .
\end{equation}
We call this variant the \emph{Poisson Mixed-Topic Link Model with Degree Correction} (\MTLMDC).


\begin{figure}
	\centering
	\subfigure[\LINKLDA]{
        \epsfig{file=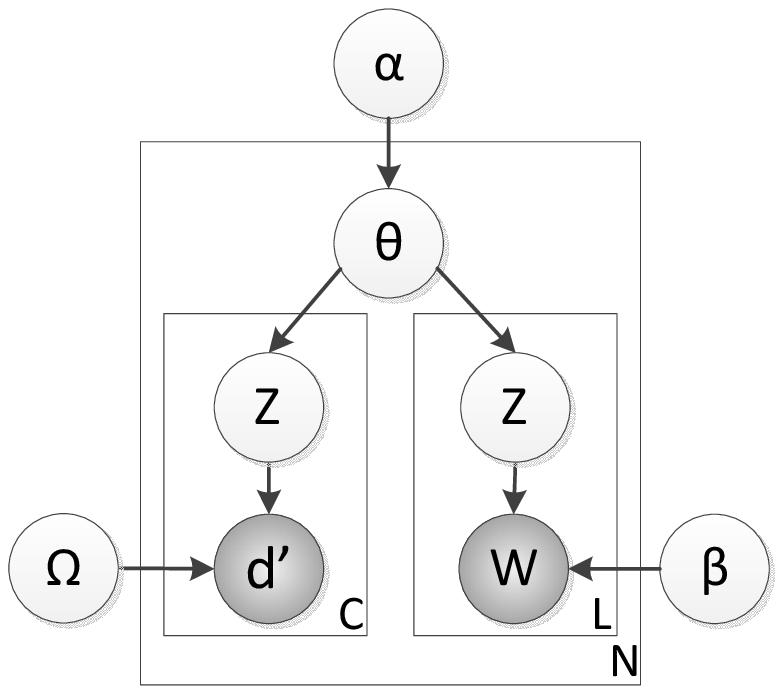,width=0.2\textwidth}
	    \label{fig:gm_linklda}
	}
	\subfigure[\CPLDC]{
        \epsfig{file=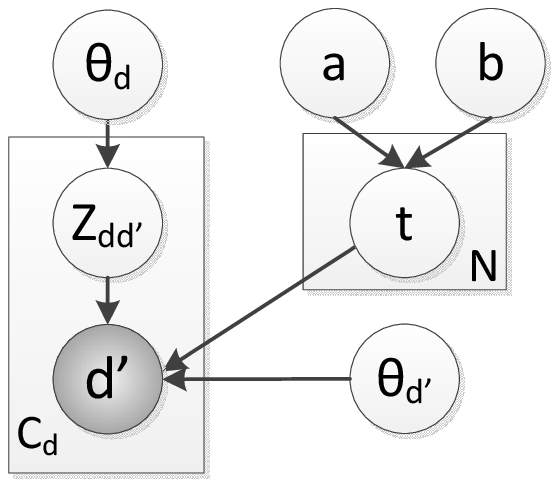,width=0.15\textwidth}
	    \label{fig:gm_cpldc}
	}
	\subfigure[\RTM]{
        \epsfig{file=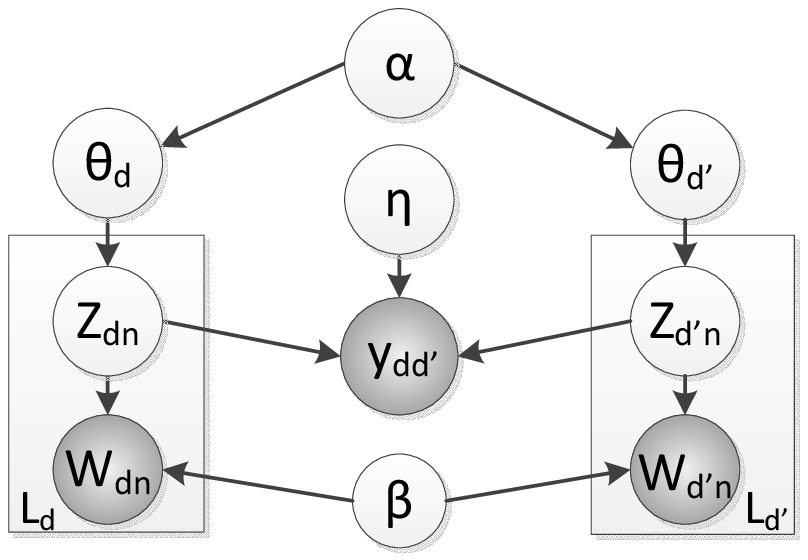,width=0.2\textwidth}
	    \label{fig:gm_rtm}
	}
	\subfigure[\MTLMDC]{
        \epsfig{file=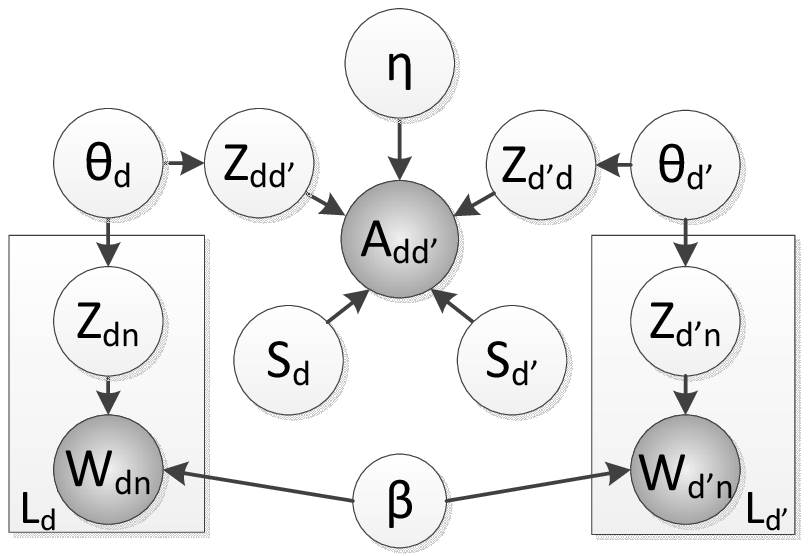,width=0.2\textwidth}
	    \label{fig:gm_mtlmdc}
	}
	\caption[Optional caption for list of figures]{\label{fig:GM} Graphical models for link generation.}
\end{figure}

\subsection{Prior Work on Content--Link Models}
\label{sec:prior}


Most models for document networks generate content using either \PLSA~\cite{Hofmann1999}, as we do, or \LDA~\cite{Blei2003}.  The distinction is that \PLSA\ treats the document mixtures $\theta_d$ as parameters, while in \LDA\ they are hidden variables, integrated over a Dirichlet distribution.  As we show in~\secref{sec:algorithm}, our approach gives a simple, exact EM algorithm, avoiding the need for sampling or variational methods.  While we do not impose a Dirichlet prior on $\theta_d$ in this paper, it is easy to add a corresponding term to the update equations for the EM algorithm,
with no loss of efficiency.

There are a variety of methods in the literature to generate links between documents.  \PHITSPLSA~\cite{Cohn2001}, \LINKLDA~\cite{Erosheva2004} and \LINKPLSALDA~\cite{Nallapati2008} use the \PHITS~\cite{cohn2000} model for link generation. \PHITS\ treats each document as an additional term in the vocabulary, so two documents are similar if they link to the same documents.  This is analogous to a mixture model for networks studied in~\cite{NewmanLeicht}.  In contrast, block models like ours treat documents as similar if they link to \emph{similar} documents, as opposed to literally the same ones.

The \PAIRLINKLDA\ model~\cite{Nallapati2008}, like ours, generates the links with a mixed-topic block model, although as in \MMSBM~\cite{Airoldi2008} and \LDA~\cite{Blei2003} it treats the $\theta_d$ as hidden variables integrated over a Dirichlet prior.  They fit their model with a variational method that requires $N^2$  parameters, making it less scalable than our approach.

In the \CPLDC\ model~\cite{Yang2009}, the link probability from $d$ to $d'$ is determined by their topic mixtures $\theta_d,\theta_{d'}$ and the popularity $t_{d'}$ of $d'$, which is drawn from a Gamma distribution with hyperparameters $a$ and $b$.  Thus $t_{d'}$ plays a role similar to the degree-correcting parameter $S_{d'}$ in our model, although we correct for the degree of $d$ as well.  However, \CPLDC\ does not generate the content, but takes it as given.

The Relational Topic Model (\RTM)~\cite{Chang2009,Chang2010} assumes that the link probability between $d$ and $d'$ depends on the topics of the words appearing in their text.  In contrast, our model uses the underlying topic mixtures $\theta_d$ to generate both the content and the links.  Like our model, \RTM\ defines the similarity of two topics as a weighted inner product of their topic mixtures: however, in \RTM\ the probability of a link is a nonlinear function of this similarity, which can be logistic, exponential or normal, of this similarity.

Although it deals with a slightly different kind of dataset, our model is closest in spirit to the Latent Topic Hypertext Model (\LTHM)~\cite{Gruber2008}.  This is a generative model for hypertext networks, where each link from $d$ to $d'$ is associated with a specific word $w$ in $d$. If we sum over all words in $d$, the total number of links $A_{dd'}$ from $d$ to $d'$ that \LTHM\ would generate follows a binomial distribution
\begin{equation}
A_{dd'} \sim \Bin\left( L_d, \lambda_{d'} \sum_z \theta_{dz} \theta_{d'z} \right) \, ,
\end{equation}
where $\lambda_{d'}$ is, in our terms, a degree-correction parameter.  When $L_d$ is large this becomes a Poisson distribution with mean $L_d \lambda_{d'} \sum_z \theta_{dz} \theta_{d'z}$.  Our model differs from this in two ways:
our parameters $\eta_z$ give a link density associated with each topic $z$, and our degree correction $S_d$ does not assume that the number of links from $d$ is proportional to its length.

We briefly mention several other approaches.  The authors of~\cite{GetoorJMLR02} extend the probabilistic relational model (\mbox{\textsc{prm}}) framework and proposed a unified generative model for both content and links in a relational structure. In \cite{luicmlws03}, the authors proposed a link-based model that describes both node attributes and links.
The \HTM\ model~\cite{Sun2008} treats links as fixed rather than generating them, and only generates the text.  Finally, the \mbox{\textsc{lmmg}} model~\cite{Lescovec-ICML12} treats the appearance or absence of a word as a binary attribute of each document, and uses a logistic or exponential function of these attributes to determine the link probabilities.

In~\secref{sec:experiments} below, we compare our model to \PHITSPLSA, \LINKLDA, \CPLDC, and \RTM.  Graphical models for the link generation components of these models, and ours, are shown in~\figref{fig:GM}.

\section{A Scalable EM Algorithm}
\label{sec:algorithm}

Here we describe an efficient Expectation-Maximization algorithm to find the maximum-likelihood estimates of the parameters of our model.  Each update takes $O(K(N+M+R))$ time for a document network with $K$ topics, $N$ documents, and $M$ links, where $R$ is the sum over the documents of the number of distinct words in each one.  Thus the running time per iteration is linear in the size of the corpus.

For simplicity we describe the algorithm for the simpler version of our model, \MTLM.  The algorithm for the degree-corrected version, \MTLMDC, is similar.

\subsection{The likelihood}

Let $C_{dw}$ denote the number of times a word $w$ appears in document $d$.  From~\eqref{eq:word_gen_prob}, the log-likelihood of $d$'s content is
\begin{align}
\Lcontent_d
= \log P(w_{d1}, \ldots, w_{dL_d} \,|\, \theta_d, \beta) \nonumber \\
= \sum_{w=1}^W C_{dw} \log \left(\sum_{z=1}^K \theta_{dz} \beta_{zw} \right) \, .
\label{eq:lcontent}
\end{align}
Similarly, from \eqref{eq:link_poission_vanilla}, the log-likelihood for the links $A_{dd'}$ is
\begin{align}
\Llinks
&= \log P(A \,|\, \theta, \eta) \nonumber \\
&= \frac{1}{2} \sum_{dd'} A_{dd'} \log \left(\sum_z\theta_{dz} \theta_{d'z} \eta_z \right) \nonumber \\
&- \frac{1}{2} \sum_{dd'} \sum_z\theta_{dz} \theta_{d'z} \eta_z
 \, .
\end{align}
We ignore the constant term $-\sum_{dd'} \log A_{dd'}!$ from the denominator of the Poisson distribution, since it has no bearing on the parameters.

\subsection{Balancing Content and Links}
\label{sec:balance}

While we can use the total likelihood $\sum_d \Lcontent_d + \Llinks$ directly, in practice we can improve our performance significantly by better balancing the information in the content vs.\ that in the links.  In particular, the log-likelihood $\Lcontent_d$ of each document is proportional to its length, while its contribution to $\Llinks$ is proportional to its degree.  Since a typical document has many more words than links, $\Lcontent$ tends to be much larger than $\Llinks$.

Following~\cite{Hofmann1999}, we can provide this balance in two ways.  One is to normalize $\Lcontent$ by the length $L_d$, and another is to add a parameter $\alpha$ that reweights the relative contributions of the two terms $\Lcontent$ and $\Llinks$.  We then maximize the function
\begin{equation}
\label{eq:weight_norm_llh}
\mathcal{L}
= \alpha \sum_d \frac{1}{L_d} \,\Lcontent_d + (1-\alpha) \Llinks \, .
\end{equation}
Varying $\alpha$ from $0$ to $1$ lets us interpolate between two extremes: studying the document network purely in terms of its topology, or purely in terms of the documents' content.  Indeed, we will see in~\secref{sec:experiments} that the optimal value of $\alpha$ depends on which task we are performing: closer to $0$ for link prediction, and closer to $1$ for topic classification.

\subsection{Update Equations and Running Time}

We maximize $\mathcal{L}$ as a function of $\{ \theta, \beta, \eta \}$ using an \EM\ algorithm, very similar to the one introduced by~\cite{Ball2011} for overlapping community detection.
We start with a standard trick to change the log of a sum into a sum of logs, writing
\begin{align}
\label{eq:llh_jensen}
\Lcontent_d &\ge
\sum_{w=1}^W C_{dw} \sum_{z=1}^K h_{dw}(z) \log \frac{\theta_{dz} \beta_{zw}}{h_{dw}(z)} \nonumber \\
\Llinks &\ge
 \frac{1}{2} \sum_{dd'} \sum_{z=1}^K A_{dd'} q_{dd'}(z) \log \frac{\theta_{dz} \theta_{d'z} \eta_z}{q_{dd'}(z)} \nonumber \\
&- \frac{1}{2} \sum_{dd'} \sum_{z=1}^K \theta_{dz} \theta_{d'z} \eta_z
 \, .
\end{align}
Here $h_{dw}(z)$ is the probability that a given appearance of $w$ in $d$ is due to topic $z$, and $q_{dd'}(z)$ is the probability that a given link from $d$ and $d'$ is due to topic $z$.  This lower bound holds with equality when
\begin{align}
\label{eq:hqdist_update}
 h_{dw}(z) = \frac{\theta_{dz} \beta_{zw}}{\sum_{z'} \theta_{dz'} \beta_{z'w}}
\, , \;
 q_{dd'}(z) = \frac{\theta_{dz} \theta_{d'z} \eta_z}{\sum_{z'} \theta_{dz'} \theta_{d'z'} \eta_{z'}} \, ,
\end{align}
giving us the E step of the algorithm.

For the M step, we derive update equations for the parameters $\{\theta,\beta,\eta\}$.  By taking derivatives of the log-likelihood~\eqref{eq:weight_norm_llh} (see the \appref{sec:appedix_a} for details) we obtain
\begin{gather}
\label{eq:mle_eta}
 \eta_z
 = \frac{\sum_{dd'} A_{dd'} q_{dd'}(z)}{\left( \sum_d \theta_{dz} \right)^2} \\
 \label{eq:mle_beta}
 \beta_{zw}
= \frac{\sum_d (1/L_d) \,C_{dw} h_{dw}(z)}{\sum_d (1/L_d) \sum_{w'} C_{dw'} h_{dw'}(z)} \\
\label{eq:mle_theta}
 \theta_{dz}
 = \frac{(\alpha / L_d) \sum_w C_{dw} h_{dw}(z) + (1-\alpha) \sum_{d'} A_{dd'} q_{dd'}(z)}{\alpha+(1-\alpha)\kappa_d} \, .
\end{gather}
Here $\kappa_d = \sum_{d'} A_{dd'}$ is the degree of document $d$.

To analyze the running time, let $R_d$ denote the number of distinct words in document $d$, and let $R = \sum_d R_d$.  Then only $KR$ of the parameters $h_{dw}(z)$ are nonzero.  Similarly, $q_{dd'}(z)$ only appears if $A_{dd'} \ne 0$, so in a network with $M$ links only $KM$ of the $q_{dd'}(z)$ are nonzero.  The total number of nonzero terms appearing in~\eqref{eq:hqdist_update}--\eqref{eq:mle_theta}, and hence the running time of the E and M steps, is thus $O(K(N+M+R))$.

As in~\cite{Ball2011}, we can speed up the algorithm if $\theta$ is sparse, i.e. if many documents belong to fewer than $K$ topics, so that many of the $\theta_{dz}$ are zero.
According to~\eqref{eq:hqdist_update}, if $\theta_{dz}=0$ then $h_{d\ell}(z) = q_{dd'}(z) = 0$, in which case~\eqref{eq:mle_theta} implies that $\theta_{dz}=0$ for all future iterations.  If we choose a threshold below which $\theta_{dz}$ is effectively zero, then as $\theta$ becomes sparser we can maintain just those $h_{d\ell}(z)$ and $q_{dd'}(z)$ where $\theta_{dz} \ne 0$.  This in turn simplifies the updates for $\eta$ and $\beta$ in~\eqref{eq:mle_eta} and~\eqref{eq:mle_beta}.

%

We note that the simplicity of our update equations comes from the fact that the $A_{dd'}$ is Poisson, and that its mean is a multilinear function of the parameters.  Models where $A_{dd'}$ is Bernoulli-distributed with a more complicated link probability, such as a logistic function, have more complicated derivatives of the likelihood, and therefore more complicated update equations.

Note also that this EM algorithm is exact, in the sense that the maximum-likelihood estimators $\{ \hattheta, \hatbeta, \hateta \}$ are fixed points of the update equations.  This is because the E step~\eqref{eq:hqdist_update} is exact, since the conditional distribution of topics associated with each word occurrence and each link is a product distribution, which we can describe exactly with $h_{dw}$ and $q_{dd'}$.  (There are typically multiple fixed points, so in practice we run our algorithm with many different initial conditions, and take the fixed point with the highest likelihood.)

This exactness is due to the fact that the topic mixtures $\theta_d$ are parameters to be inferred.  In models such as \LDA\ and \MMSBM\ where $\theta_d$ is a hidden variable integrated over a Dirichlet prior, the topics associated with each word and link have a complicated joint distribution that can only be approximated using sampling or variational methods.  (To be fair, recent advances such as stochastic optimization based on network subsampling~\cite{gopalan2012} have shown that  approximate inference in these models can be carried out quite efficiently.)

On the other hand, in the context of finding communities in networks, models with Dirichlet priors have been observed to generalize more successfully than Poisson models such as \BKN~\cite{gopalan2012}.  Happily, we can impose a Dirichlet prior on $\theta_d$ with no loss of efficiency, simply by including pseudocounts in the update equations---in essence adding additional words and links that are known to come from each topic (see~\appref{sec:appedix_c}).  This lets us obtain a maximum a posteriori (MAP) estimate of an \LDA-like model.  We leave this as a direction for future work.

\subsection{Discrete Labels and Local Search}
\label{sec:discrete-labels}

Our model, like \PLSA\ and the \BKN\ model, lets us infer a soft classification---a mixture of topic labels or community memberships for each document.  However, we often want to infer categorical labels, where each document $d$ is assigned to a single topic $1 \le z_d \le K$.  A natural way to do this is to let $z_d$ be the most-likely label in the inferred mixture, $\hat{z}_d = \argmax_z \theta_{dz}$.
This is equivalent to rounding $\theta_d$ to a delta function, $\theta_{dz}=1$ for $z=\hat{z}_d$ and $0$ for $z \ne \hat{z}_d$.

If we wish, we can improve these discrete labels further using local search.  If each document has just a single topic, the log-likelihood of our model is
\begin{align}
\label{eq:non-overlapping-lc}
\Lcontent_d  &= \sum_{w=1}^{W}C_{dw} \log \beta_{z_d w} \\
\label{eq:non-overlapping-ll}
\Llinks &= \frac{1}{2} \sum_{dd'} A_{dd'} \log \eta_{z_d z_{d'}} \, .
\end{align}
Note that here $\eta$ is a matrix, with off-diagonal entries that allow documents with different topics $z_d, z_{d'}$to be linked.  Otherwise, these discrete labels would cause the network to split into $K$ separate components.

Let $n_z$ denote the number of documents of topic $z$, let $L_z = \sum_{d: z_d=z} L_d$ be their total length, and let $C_{zw} = \sum_{d: z_d=z} C_{dw}$ be the total number of times $w$ appears in them.  Let $m_{zz'}$ denote the total number of links between documents of topics $z$ and $z'$, counting each link twice if $z=z'$.  Then the MLEs for $\beta$ and $\eta$ are
\begin{align}
 \hat{\beta}_{zw} = \frac{C_{zw}}{L_z}
 \, , \;
 \hat{\eta}_{zz'} = \frac{m_{zz'}}{n_z n_{z'}} \, .
\end{align}
Applying these MLEs in~\eqref{eq:non-overlapping-lc} and~\eqref{eq:non-overlapping-ll} gives us a point estimate of the likelihood of a discrete topic assignment $z_d$, which we can normalize or reweight as discussed in~\secref{sec:balance} if we like.  We can then maximize this likelihood using local search: for instance, using the Kernighan-Lin heuristic as in~\cite{Karrer2011} or a Monte Carlo algorithm to find a local maximum of the likelihood in the vicinity of $\hat{z}$.
Each step of these algorithms changes the label of a single document $d$, so we can update the values of $n_z$, $L_z$, $C_{zw}$, and $m_{zz'}$ and compute the new likelihood in $O(K+R_d)$ time.  In our experiments we used the KL heuristic, and found that for some data sets it noticeably improved the accuracy of our algorithm for the document classification task.

\section{Experimental Results}
\label{sec:experiments}

In this section we present empirical results on our model and our algorithm for unsupervised document classification and link prediction.  We compare its accuracy and running time with those of several other methods, testing it on three real-world document citation networks.

\subsection{Data Sets}

The top portion of~\tabref{tab:scalability_test} lists the basic statistics for three real-world corpora~\cite{Sen2008}: Cora, Citeseer, and PubMed\footnote{These data sets are available for download at \url{http://www.cs.umd.edu/projects/linqs/projects/lbc/}}.  Cora and Citeseer contain papers in machine learning, with $K=7$ topics for Cora and $K=6$ for Citeseer.  PubMed consists of medical research papers on $K=3$ topics, namely three types of diabetes.  All three corpora have ground-truth topic labels provided by human curators.

The data sets for these corpora are slightly different.  The PubMed data set has the number of times $C_{dw}$ each word appeared in each document, while the data for Cora and Citeseer records whether or not a word occurred at least once in the document.  For Cora and Citeseer, we treat $C_{dw}$ as being $0$ or $1$.

\subsection{Models and Implementations}
We compare the Poisson Mixed-Topic Link Model (\MTLM) and its degree-corrected variant, denoted \MTLMDC, with \PHITSPLSA, \LINKLDA, \CPLDC, and \RTM\ (see~\secref{sec:prior}).  We used our own implementation of both \PHITSPLSA\ and \RTM.
For \RTM, we implemented the variational EM algorithm given in~\cite{Chang2010}. The implementation is based on the \LDA\ code available from the authors\footnote{See \url{http://www.cs.princeton.edu/~blei/lda-c/}}. We also tried the code provided by J. Chang\footnote{See \url{http://www.cs.princeton.edu/~blei/lda/}}, which uses a Monte Carlo algorithm for the E step, but we found the variational algorithm works better on our data sets. While \RTM\ includes a variety of link probability functions, we only used the sigmoid function. We also assume a symmetric Dirichlet prior. The results for \LINKLDA\ and \CPLDC\ are taken from~\cite{Yang2009}.

Each E and M step of the variational algorithm for \RTM\ performs multiple iterations until they converge on estimates for the posterior and the parameters~\cite{Chang2010}. This is quite different from our EM algorithm: since our E step is exact, we update the parameters only once in each iteration. In our implementation, the convergence condition for the E step and for the entire EM algorithm are that the fractional increase of the log-likelihood between iterations is less than $10^{-6}$; we performed a maximum of $500$ iterations of the \RTM\ algorithm due to its greater running time. In order to optimize the $\eta$ parameters (see the graphical model in \secref{sec:prior}) \RTM\ uses a tunable regularization parameter $\rho$, which can be thought of as the number of  observed non-links.  We tried various settings for $\rho$, namely $0.1M, 0.2M, 0.5M, M, 2M, 5M$ and $10M$ where $M$ is the number of observed links.


As described in~\secref{sec:balance}, for \MTLM, \MTLMDC\ and \PHITSPLSA\ we vary the relative weight $\alpha$ of the likelihood of the content vs.\ the links, tuning $\alpha$ to its best possible value.  For the PubMed data set, we also normalized the content likelihood by the length of the documents.

\subsection{Document Classification}
\label{sec:doc_classification}
\subsubsection{Experimental Setting}
\begin{figure}
\centering
\epsfig{file=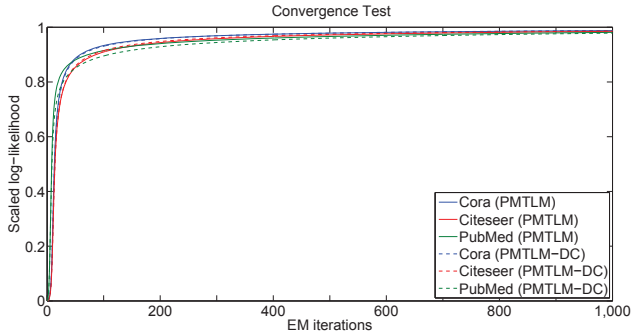,width=0.99 \columnwidth}
\caption{The log-likelihood of the \textsc{PMTLM} and \textsc{PMTLM-DC} models as a function of the number of EM iterations, normalized so that $0$ and $1$ are the initial and final log-likelihood respectively.  The convergence is roughly the same for all three data sets, showing that the number of iterations is roughly constant as a function of the size of the corpus. \label{fig:convergence_curves}}
\end{figure}

For \MTLM, \MTLMDC\ and \PHITSPLSA, we performed $500$ independent runs of the EM algorithm, each with random initial values of the parameters and topic mixtures.  For each run we iterated the EM algorithm up to $5000$ times; we found that it typically converges in fewer iterations, with the criterion that
the fractional increase of the log-likelihood for two successive iterations is less than $10^{-7}$.
\figref{fig:convergence_curves} shows that the log-likelihood as a function of the number of iterations are quite similar for all three data sets, even though these corpora have very different sizes.  This indicates that even for large data sets, our algorithm converges within a small number of iterations, making its total running time linear in the size of the corpus.

For \MTLM\ and \MTLMDC, we obtain discrete topic labels by running our EM algorithm and rounding the topic mixtures as described in~\secref{sec:discrete-labels}.  We also tested improving these labels with local search, using the Kernighan-Lin heuristic to change the label of one document at a time until we reach a local optimum of the likelihood. More precisely, of those $500$ runs, we took the $T$ best fixed points of the EM algorithm (i.e., with the highest likelihood) and attempted to improve them further with the KL heuristic.  We used
$T=50$ for Cora and Citeseer and $T=5$ for PubMed.

For \RTM, in each E step, we initialize the variational parameters randomly, and in each M step we initialize the hyperparameters randomly. We execute 500 independent runs for each setting of the tunable parameter $\rho$.

\subsubsection{Metrics}

For each algorithm, we used several measures of the accuracy of the inferred labels as compared to the human-curated ones.  The Normalized Mutual Information (\NMI) between two labelings $C_1$ and $C_2$ is defined as
\begin{align}
\NMI(C_1,C_2)=\frac{\MI(C_1,C_2)}{\textrm{max}(\entropy(C_1),\entropy(C_2))}\,.
\end{align}
Here $\MI(C_1,C_2)$ is the mutual information between $C_1$ and $C_2$, and $\entropy(C_1)$ and $\entropy(C_2)$ are the entropies of $C_1$ and $C_2$ respectively.  Thus the \NMI\ is a measure of how much information the inferred labels give us about the true ones.  We also used the Pairwise F-measure (\PWF)~\cite{basu2005} and the Variation of Information (\VI)~\cite{meilua2003} (which we wish to minimize).


\begin{table}
  \resizebox{\columnwidth}{!}{
  \centering
    \begin{tabular}{|rl|r|r|r|}
    \addlinespace
    \hline
          &       & Cora  & Citeseer & PubMed \\
    \hline
    \multicolumn{1}{|c}{\multirow{4}[8]{*}{Statistics}} & \multicolumn{1}{|l|}{$K$} & 7     & 6     & 3 \\\cline{2-5}
    \multicolumn{1}{|c}{} & \multicolumn{1}{|l|}{$N$} & 2,708 & 3,312 & 19,717 \\\cline{2-5}
    \multicolumn{1}{|c}{} & \multicolumn{1}{|l|}{$M$} & 5,429 & 4,608 & 44,335 \\\cline{2-5}
    \multicolumn{1}{|c}{} & \multicolumn{1}{|l|}{$W$} &1,433 & 3,703 & 4,209 \\\cline{2-5}
    \multicolumn{1}{|c}{} & \multicolumn{1}{|l|}{$R$} &49,216 & 105,165 & 1,333,397 \\ \hline\hline
    \multicolumn{1}{|c}{\multirow{4}[10]{*}{Time (sec)}} & \multicolumn{1}{|l|}{\EM\ (\PLSA)} & 28 & 61 & 362 \\ \cline{2-5}
    \multicolumn{1}{|c}{} & \multicolumn{1}{|l|}{\EM\ (\PHITSPLSA)} & 40 & 67 & 445 \\ \cline{2-5}
    \multicolumn{1}{|c}{} & \multicolumn{1}{|l|}{\EM\ (\MTLM)} & 33 & 64 & 419 \\ \cline{2-5}
    \multicolumn{1}{|c}{} & \multicolumn{1}{|l|}{\EM\ (\MTLMDC)} & 36 & 64 & 402 \\ \cline{2-5}
    \multicolumn{1}{|c}{} & \multicolumn{1}{|l|}{\EM\ (\RTM)} & 992 & 597 & 2,194 \\ \cline{2-5}
    \multicolumn{1}{|c}{} & \multicolumn{1}{|l|}{KL\, (\MTLM)} & 375 & 618 & 13,723 \\ \cline{2-5}
    \multicolumn{1}{|c}{} & \multicolumn{1}{|l|}{KL\, (\MTLMDC)} & 421 & 565 & 13,014 \\
    \hline
    \end{tabular}}%
      \caption{The statistics of the three data sets, and the mean running time, for the \EM\ algorithms in our model \textsc{PMTLM}, its degree-corrected variant \textsc{PMTLM-DC}, and \textsc{PLSA}, \textsc{PHITS-PLSA}, and \textsc{RTM}.  Each corpus has $K$ topics, $N$ documents, $M$ links, a vocabulary of size $W$, and a total size $R$.
Running times for our algorithm, \textsc{PLSA}, and \textsc{PHITS-PLSA} are given for one run of $5000$ \EM\ iterations.  Running times for \textsc{RTM} consist of up to 500 \EM\ iterations, or until the convergence criteria are reached.  Our \EM\ algorithm
is highly scalable, with a running time that grows linearly with the size of the corpus.  In particular, it is much faster that the variational algorithm for \textsc{RTM}.  Improving discrete labels with the Kernighan-Lin heuristic (KL) increases our algorithm's running time, but improves its accuracy for document classification in Cora and Citeseer.
    \label{tab:scalability_test}}
\end{table}%

\begin{table*}
  \resizebox{\textwidth}{!}{
  \centering
    \begin{tabular}{rrrrrrrrrr}
    \addlinespace
    \toprule
          & & Cora &  &  & Citeseer &  &  & PubMed & \\
    \midrule
    Algorithm  & \NMI   & \VI  & \PWF  & \NMI  & \VI  & \PWF  & \NMI  & \VI  & \PWF\\
    \midrule
    \PHITSPLSA\  & 0.382 (.4) & 2.285 (.4) & 0.447 (.3) & 0.366 (.5) & 2.226 (.5) & 0.480 (.5) & 0.233 (1.0) &1.633 (1.0) & 0.486 (1.0) \\
    \LINKLDA & $0.359^{\dagger}$ & --- & $0.397^{\dagger}$ & $0.192^{\dagger}$ & --- & $0.305^{\dagger}$ & --- & --- & ---\\
    \CPLDC\  & $0.489^{\dagger}$ & --- & $0.464^{\dagger}$ & $0.276^{\dagger}$ & --- & $0.361^{\dagger}$ & --- & --- & ---\\
    \RTM\  & 0.349 & 2.306 & 0.422 & 0.369 & 2.209 & 0.480 & 0.228 & 1.646 & 0.482\\
    \MTLM\ & 0.467 (.4) & 1.957 (.4) & 0.509 (.3) & 0.399 (.4) & 2.106 (.4) & 0.509 (.3) & 0.232 (.9) & 1.639 (1.0) & 0.486 (.9)\\
    \MTLM\ (\KL) & \textbf{0.514} (.4) & \textbf{1.778} (.4) &  \textbf{0.525} (.4) & \textbf{0.414} (.6) & \textbf{2.057} (.6) & 0.518 (.5) & 0.233 (.9) & 1.642 (.9) & 0.488 (.9)\\
    \MTLMDC\ & 0.474 (.3) &1.930 (.3) & 0.498 (.3) &0.402 (.3)	&2.096 (.3) & 0.518 (.3) & \textbf{0.270} (.8)	&\textbf{1.556} (.8) & \textbf{0.496} (.8)\\
    \MTLMDC\ (\KL) & 0.491 (.3) & 1.865 (.3) & 0.511 (.3) & 0.406 (.3) & 2.084 (.3) & \textbf{0.520} (.3) &0.260 (.8) & 1.577 (.8) &0.492 (.8) \\
    \bottomrule
    \end{tabular}}%
    \caption{The best normalized mutual information (\NMI), variational of information (\VI) and pairwise F-measure (\PWF) achieved by each algorithm.  Values marked by $\dagger$ are quoted from~\cite{Yang2009}; other values are based on our implementation.  The best values are shown in bold; note that we seek to maximize \NMI\ and \PWF, and minimize \VI.  For \textsc{PHITS-PLSA}, \textsc{PMTLM}, and \textsc{PMTLM-DC}, the number in parentheses is the best value of the relative weight $\alpha$ of content vs.\ links.  Refining the labeling returned by the EM algorithm with the Kernighan-Lin heuristic is indicated by \textsc(KL).
  \label{tab:cora-citeseer-pubmed}}
\end{table*}%

\subsubsection{Results}

The best \NMI, \VI, and \PWF\ we observed for each algorithm are given in~\tabref{tab:cora-citeseer-pubmed}, where for \LINKLDA\ and \CPLDC\ we quote results from~\cite{Yang2009}. For \RTM, we give these metrics for the labeling with the highest likelihood, using the best value of $\rho$ for each metric.

We see that even without the additional step of local search, our algorithm does very well, outperforming all other methods we tried on Citeseer and PubMed and all but \CPLDC\ on Cora.  (Note that we did not test \LINKLDA\ or \CPLDC\ on PubMed.)  Degree correction (\MTLMDC) improves accuracy significantly for PubMed.

Refining our labeling with the KL heuristic improved the performance of our algorithm significantly for Cora and Citeseer, giving us a higher accuracy than all the other methods we tested.  For PubMed, local search did not increase accuracy in a statistically significant way.  In fact, on some runs it decreased the accuracy slightly compared to the initial labeling $\hat{z}$ obtained from our EM algorithm; this is counterintuitive, but it shows that increasing the likelihood of a labeling in the model can decrease its accuracy.

In~\figref{fig:weight_NMI_cora_citeseer}, we show how the performance of \MTLM, \MTLMDC, and \PHITSPLSA\ varies as a function of $\alpha$, the relative weight of content vs.\ links.  Recall that at $\alpha=0$ these algorithms label documents solely on the basis of their links, while at $\alpha=1$ they only pay attention to the content.  Each point consists of the top 20 runs with that value of $\alpha$.

 For Cora and Citeseer, there is an intermediate value of $\alpha$ at which \MTLM\ and \MTLMDC\ have the best accuracy.  However, this peak is fairly broad, showing that we do not have to tune $\alpha$ very carefully.  For PubMed, where we also normalized the content information by document length, \MTLMDC\ performs best at a particular value of $\alpha$.

%



We give the running time of these algorithms, including \MTLM\ and \MTLMDC\ with and without the \KL\ heuristic, in~\tabref{tab:scalability_test}, and compare it to the running time of the other algorithms we implemented.  Our EM algorithm is much faster than the variational EM algorithm for \RTM, and is scalable in that it grows linearly with the size of the corpus.

\subsection{Link Prediction}

Link prediction (e.g.~\cite{ClausetMooreNewman,Lu_linkpred2011,zhao2013link}) is a natural generalization task in networks, and another way to measure the quality of our model and our EM algorithm.  Based on a training set consisting of a subset of the links, our goal is to rank all pairs without an observed link according to the probability of a link between them.  For our models, we rank pairs according to the expected number of links $A_{dd'}$ in the Poisson distribution, \eqref{eq:link_poission_vanilla} and~\eqref{eq:link_poission_dc}, which is monotonic in the probability that at least one link exists.

We can then predict links between those pairs where this probability exceeds some threshold.  Since we are agnostic about this threshold and about the cost of Type I vs.\ Type II errors, we follow other work in this area by defining the accuracy of our model as the AUC, i.e. the probability that a random true positive link is ranked above a random true non-link.  Equivalently, this is the area under the \emph{receiver operating characteristic} curve (ROC).  Our goal is to do better than the baseline AUC of $1/2$, corresponding to a random ranking of the pairs.


We carried out 10-fold cross-validation, in
which the links in the original graph are partitioned into 10 subsets
with equal size. For each fold, we use one subset as the test links,
and train the model using the links in the other 9 folds.
We evaluated the AUC on the held-out links and the non-links. For Cora and Citeseer, all the non-links are used. For PubMed,
we randomly chose 10\% of the non-links for comparison. We trained the
models with the same settings as those for document classification
in~\secref{sec:doc_classification}; we executed $100$
independent runs for each test.  Note that unlike the document classification task,
here we used the full topic mixtures to predict links, not just the discrete labels consisting of the most-likely topic
for each document.

Note that \MTLMDC\ assigns $S_d$ to be zero if the degree of $d$ is
zero. This makes it impossible for $d$ to have any test link with
others if its observed degree is zero in the training data. One way
to solve this is to assign a small positive value to $S_d$ even if
$d$'s degree is zero. Our approach assigns $S_d$ to be the smallest
value among those $S_{d'}$ that are non-zero:
\begin{equation}
S_d=\mbox{min}\{S_{d'}:S_{d'}>0\}\quad\textrm{if } \kappa_d=0\,.
\end{equation}


\begin{figure*}
\centering
\epsfig{file=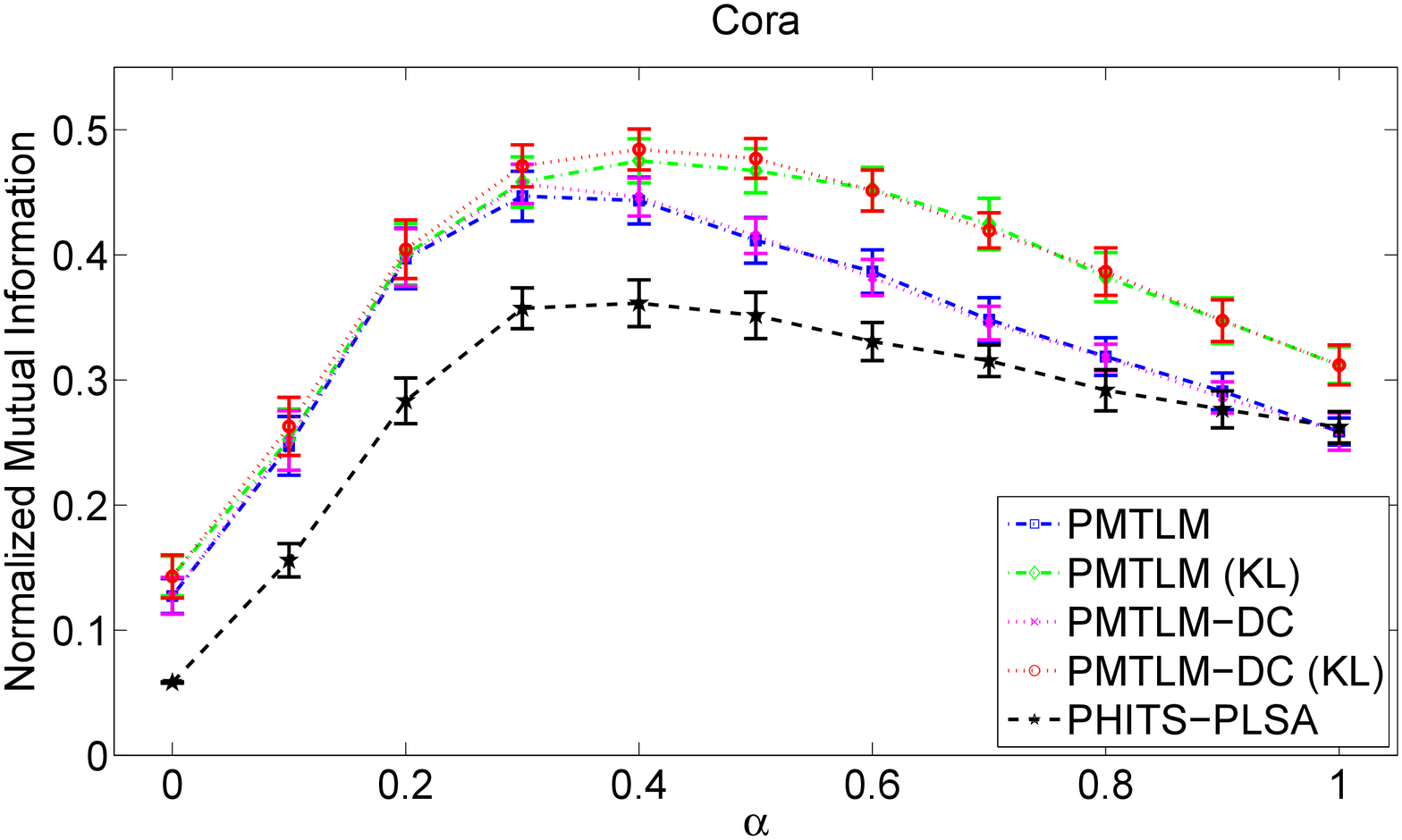,width=0.32 \textwidth, height=3.3cm}
\epsfig{file=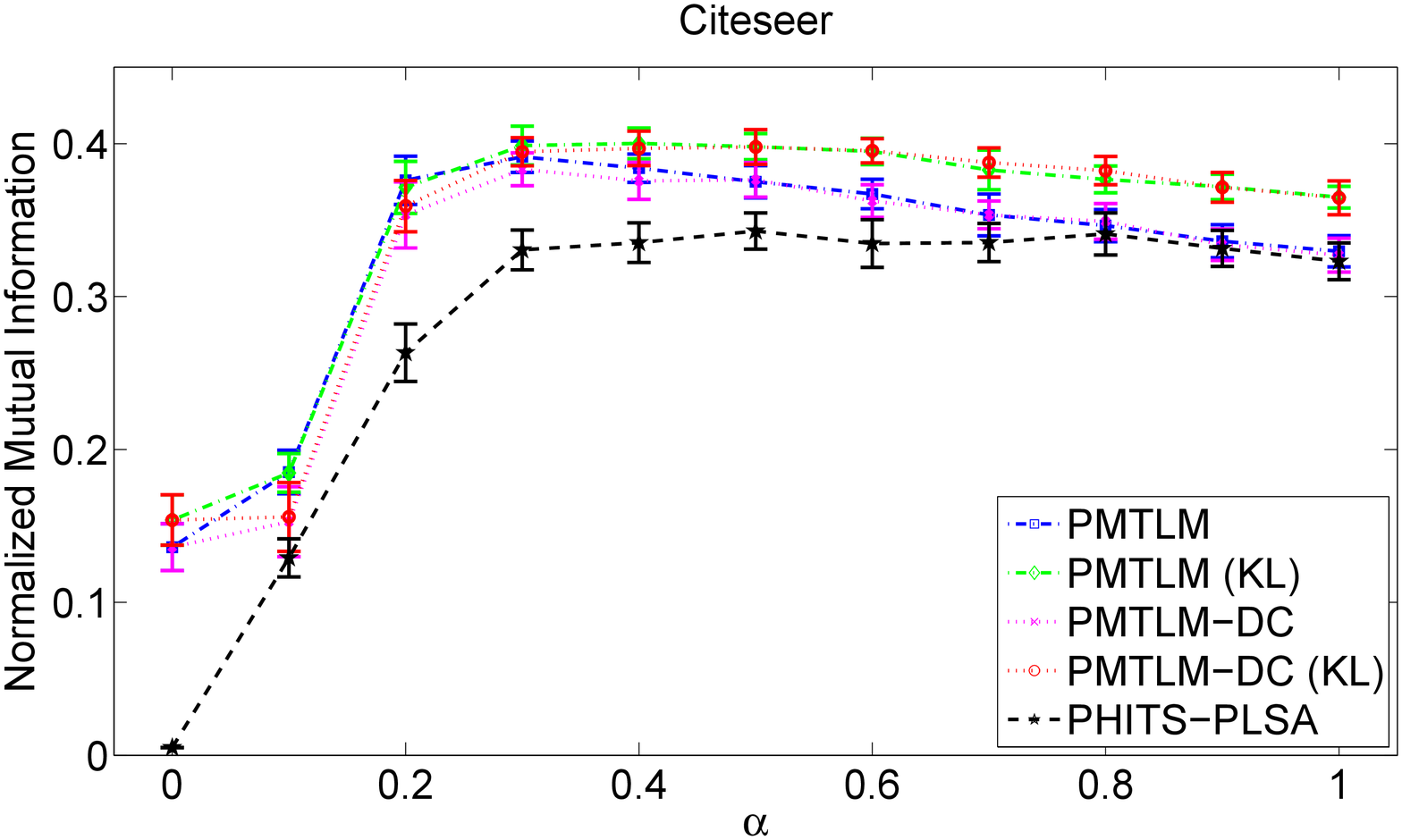,width=0.32 \textwidth, height=3.3cm}
\psfig{file=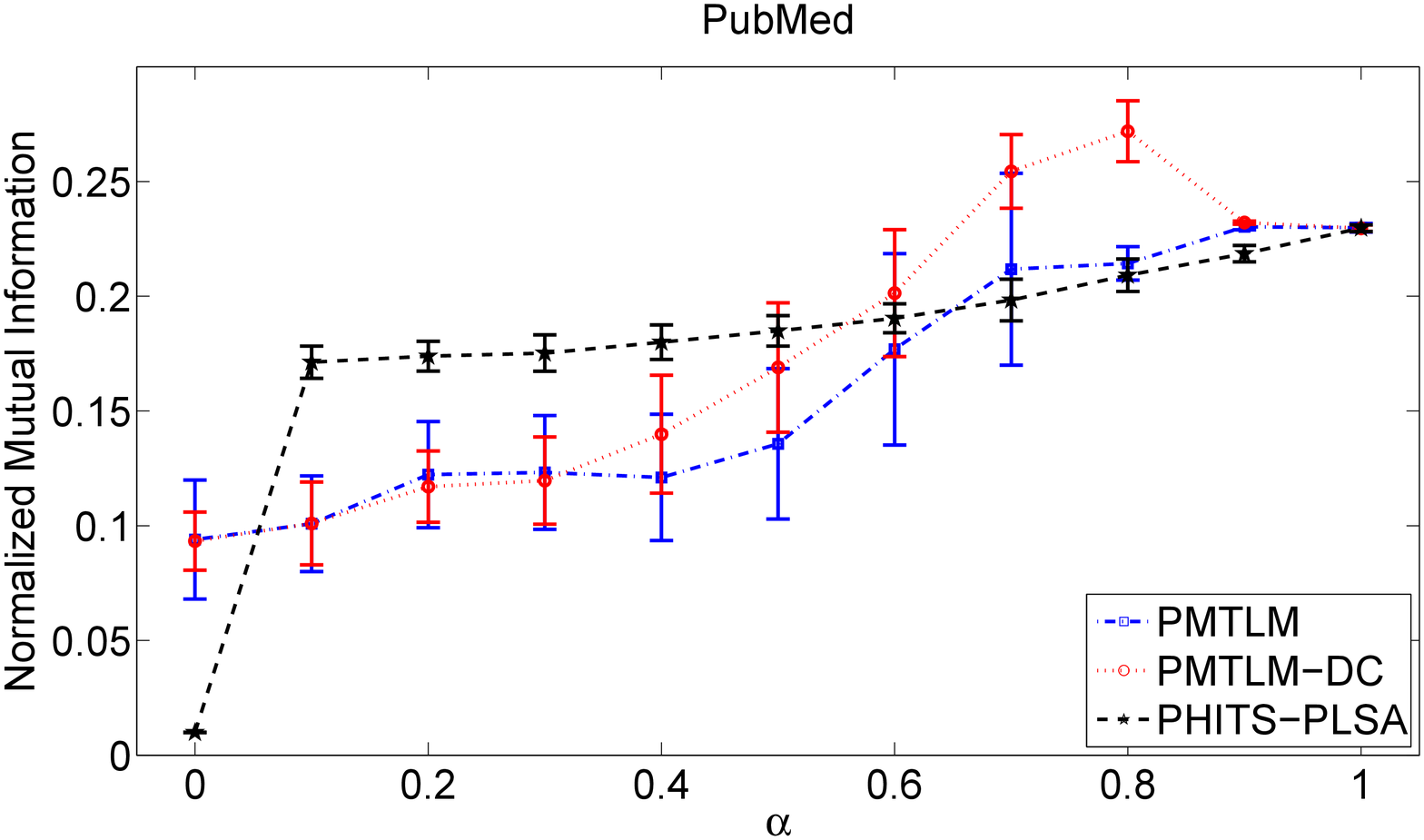,width=0.32 \textwidth, height=3.3cm}
\caption{The accuracy of \textsc{PMTLM}, \textsc{PMTLM-DC}, and \textsc{PHITS-PLSA} on the document classification task, measured by the \NMI, as a function of the relative weight $\alpha$ of the content vs.\ the links.  At $\alpha=0$ these algorithms label documents solely on the basis of their links, while at $\alpha=1$ they pay attention only to the content.  For Cora and Citeseer, there is a broad range of $\alpha$ that maximizes the accuracy.  For PubMed, the degree-corrected model \textsc{PMTLM-DC} performs best at a particular value of $\alpha$.
\label{fig:weight_NMI_cora_citeseer}}
\end{figure*}

\begin{figure*}
\centering
\subfigure[AUC values for different $\alpha$.]{
\epsfig{file=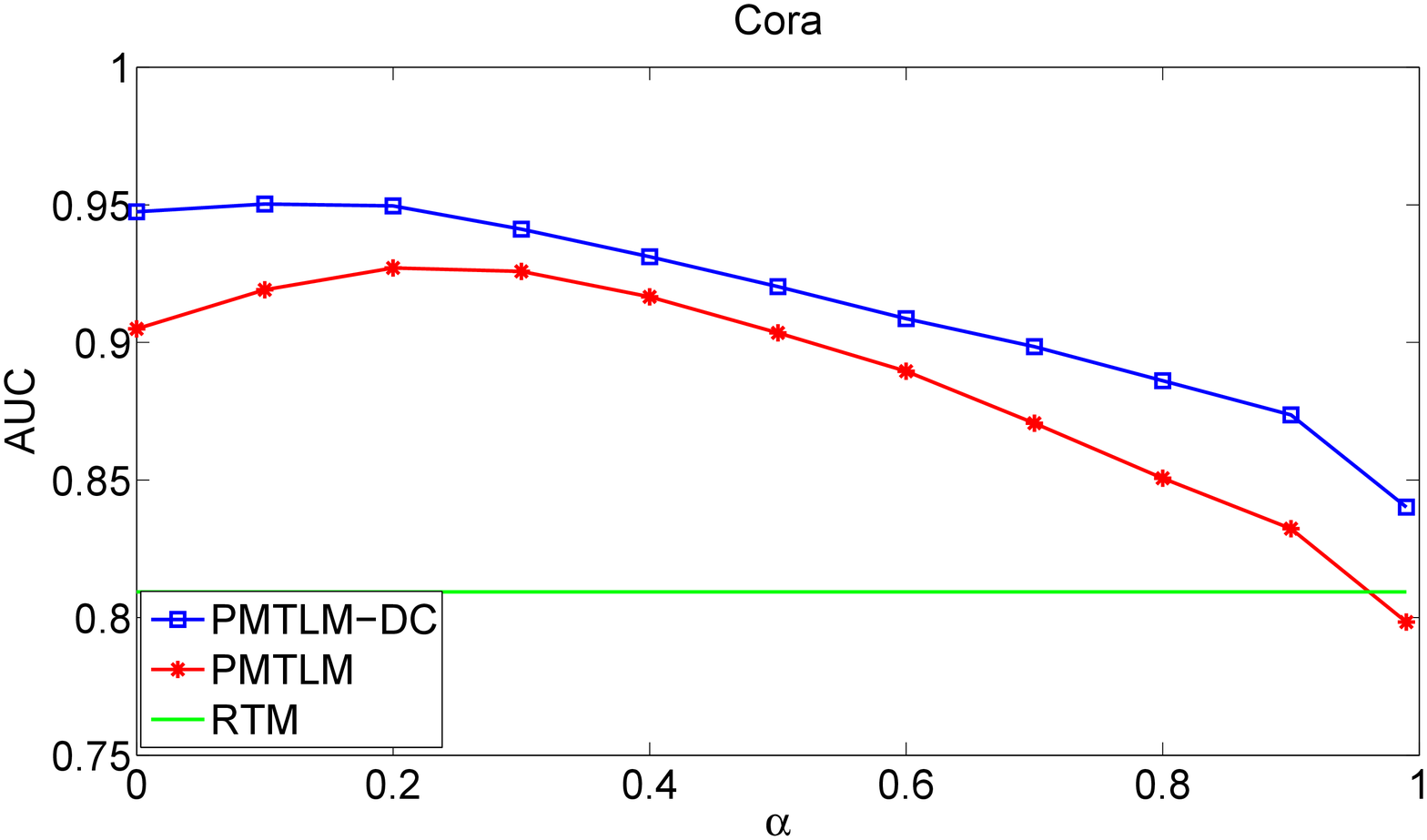,width=0.32 \textwidth, height=3.3cm}
\epsfig{file=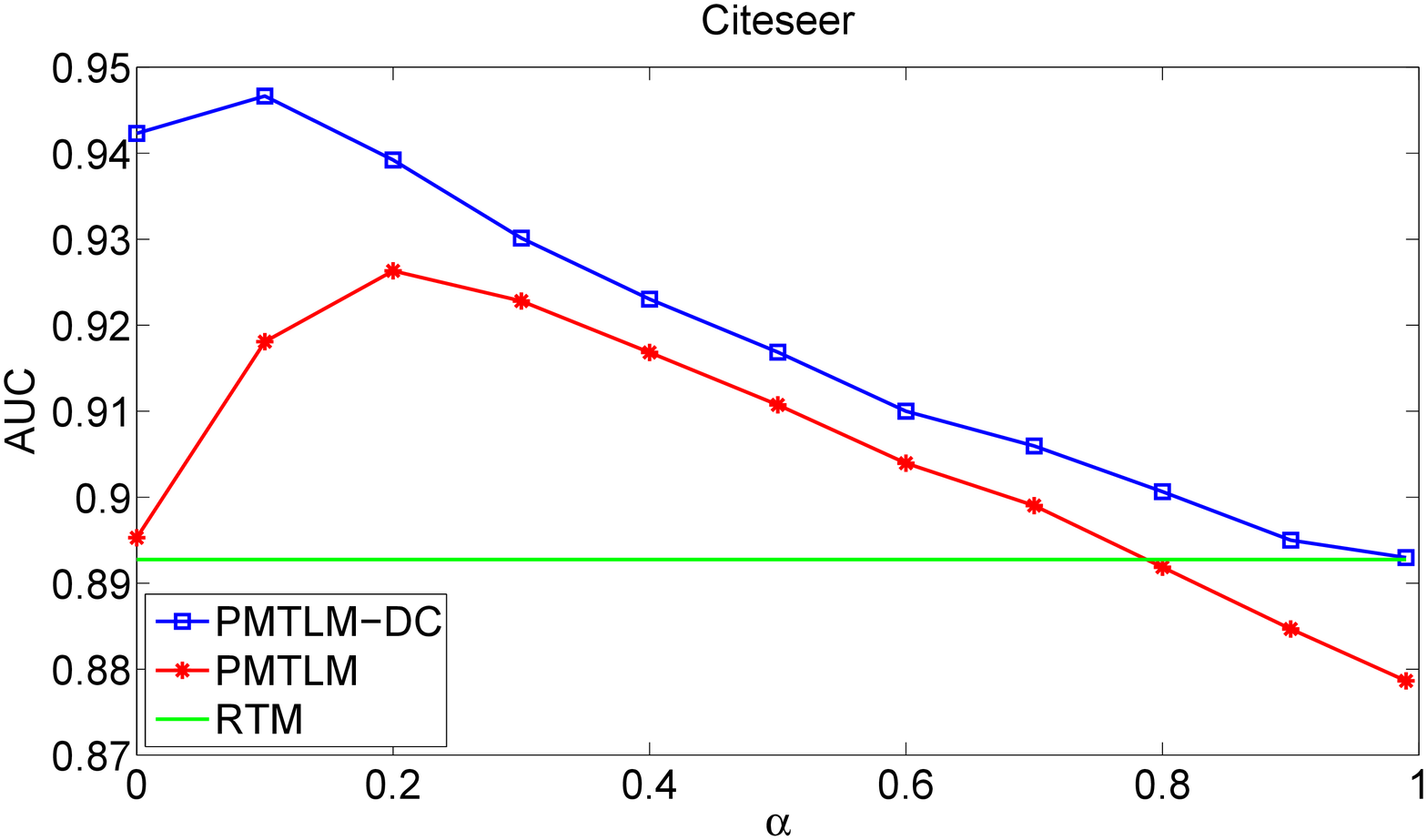,width=0.32 \textwidth, height=3.3cm}
\epsfig{file=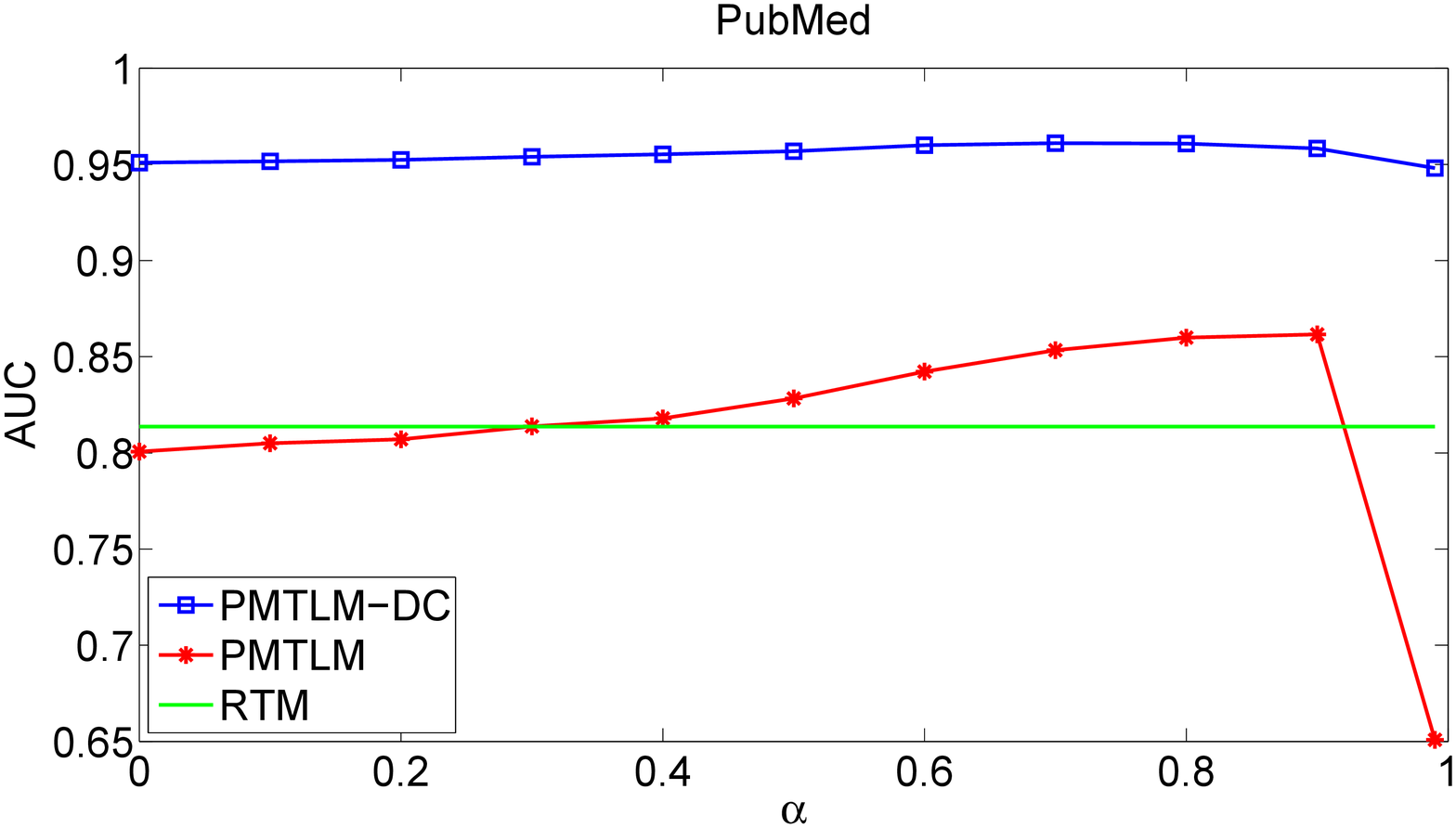,width=0.32 \textwidth, height=3.3cm}
\label{fig:AUC}
}
\subfigure[ROC curves achieving the highest AUC values.]{
\epsfig{file=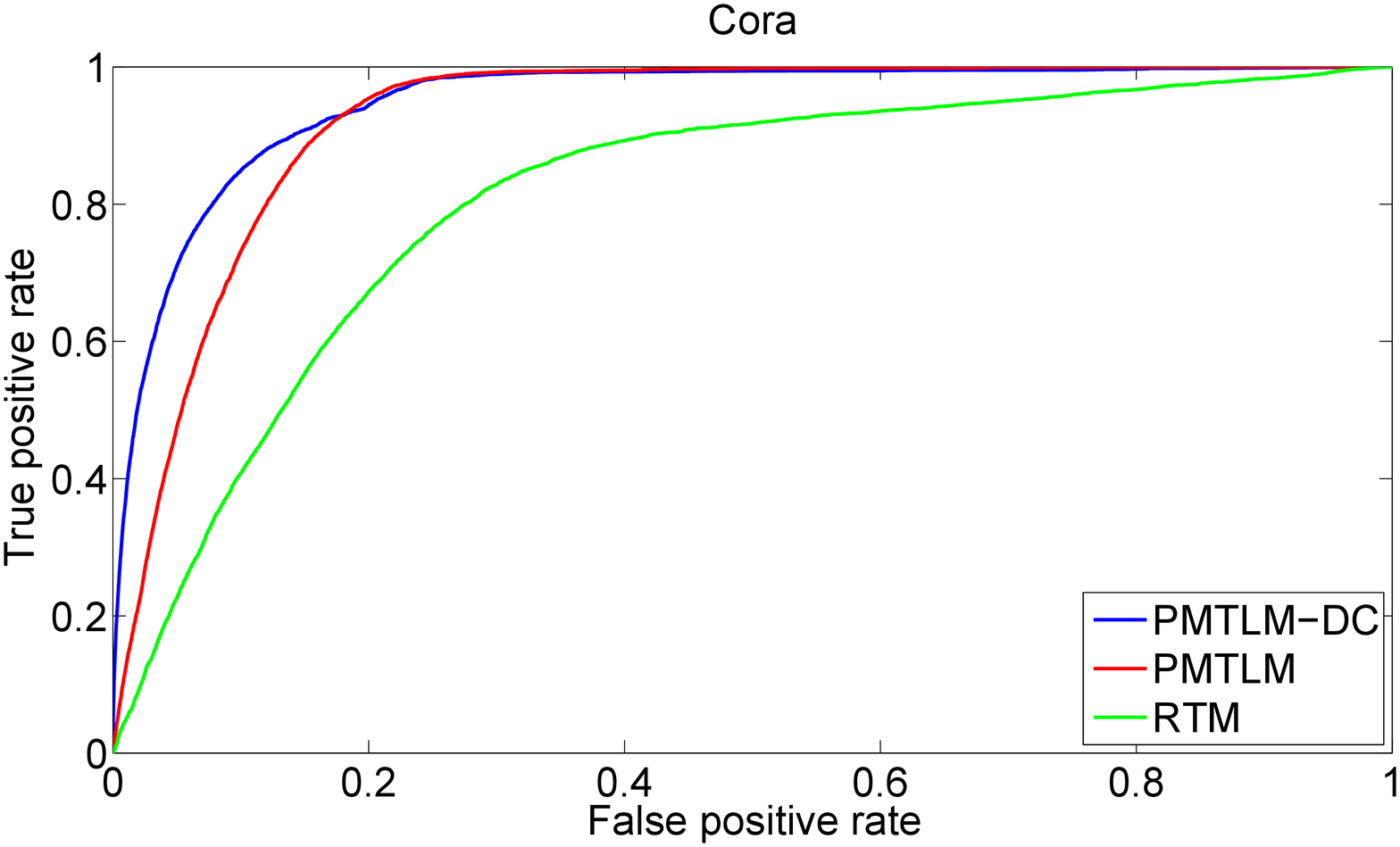,width=0.32 \textwidth, height=3.3cm}
\epsfig{file=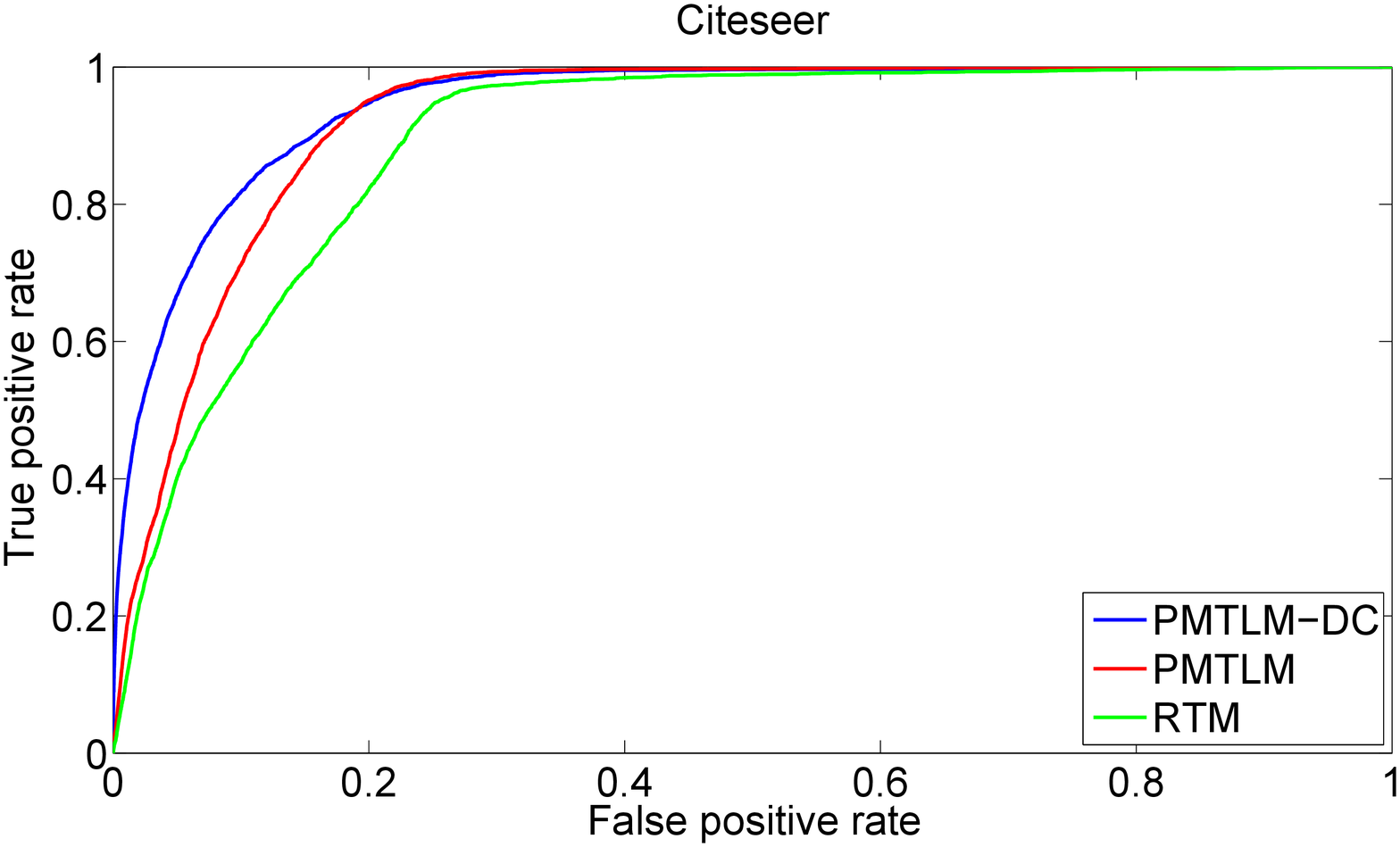,width=0.32 \textwidth, height=3.3cm}
\epsfig{file=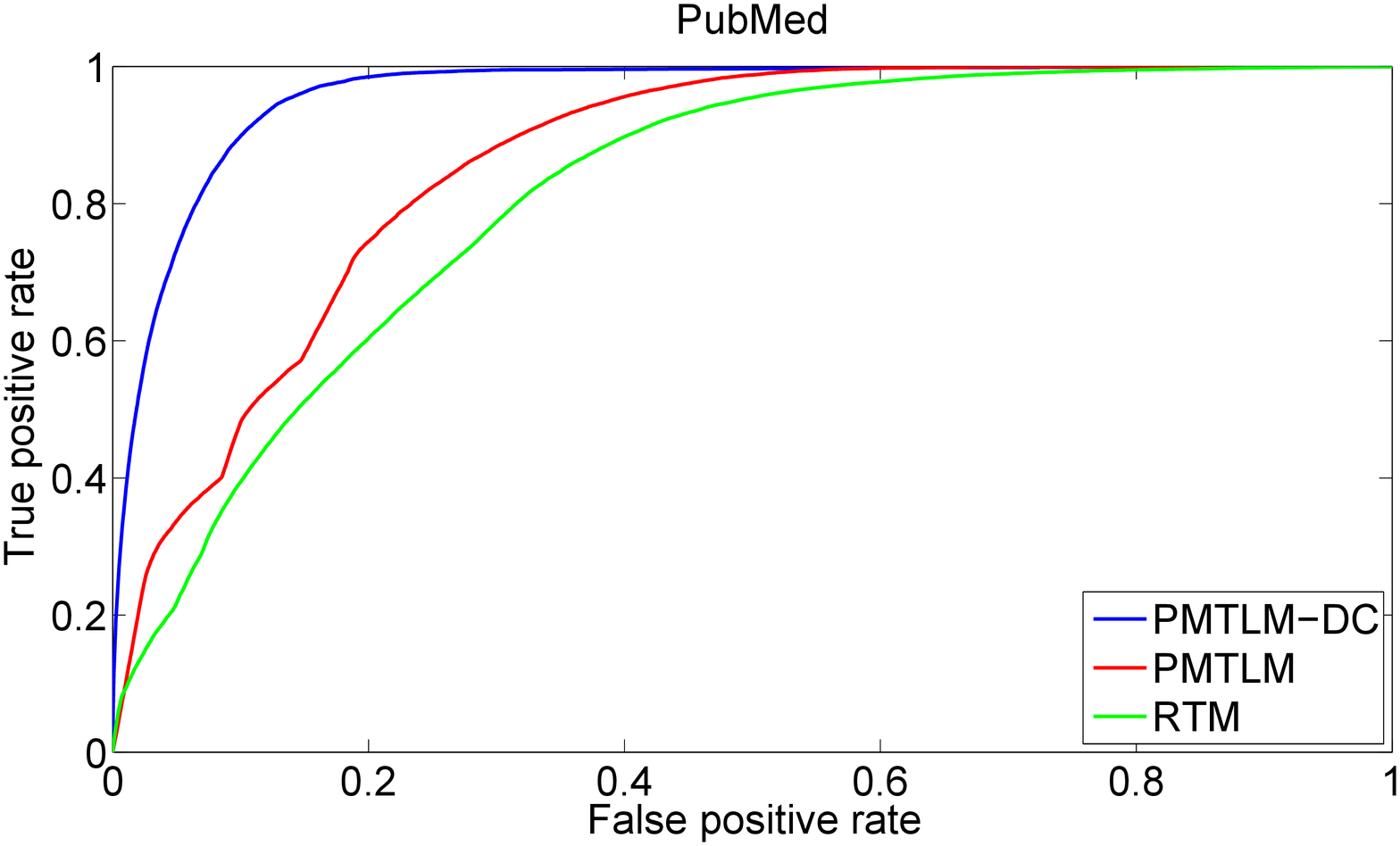,width=0.32 \textwidth, height=3.3cm}
\label{fig:ROC}
}
\subfigure[Precision-recall curves achieving the highest AUC values.]{
\epsfig{file=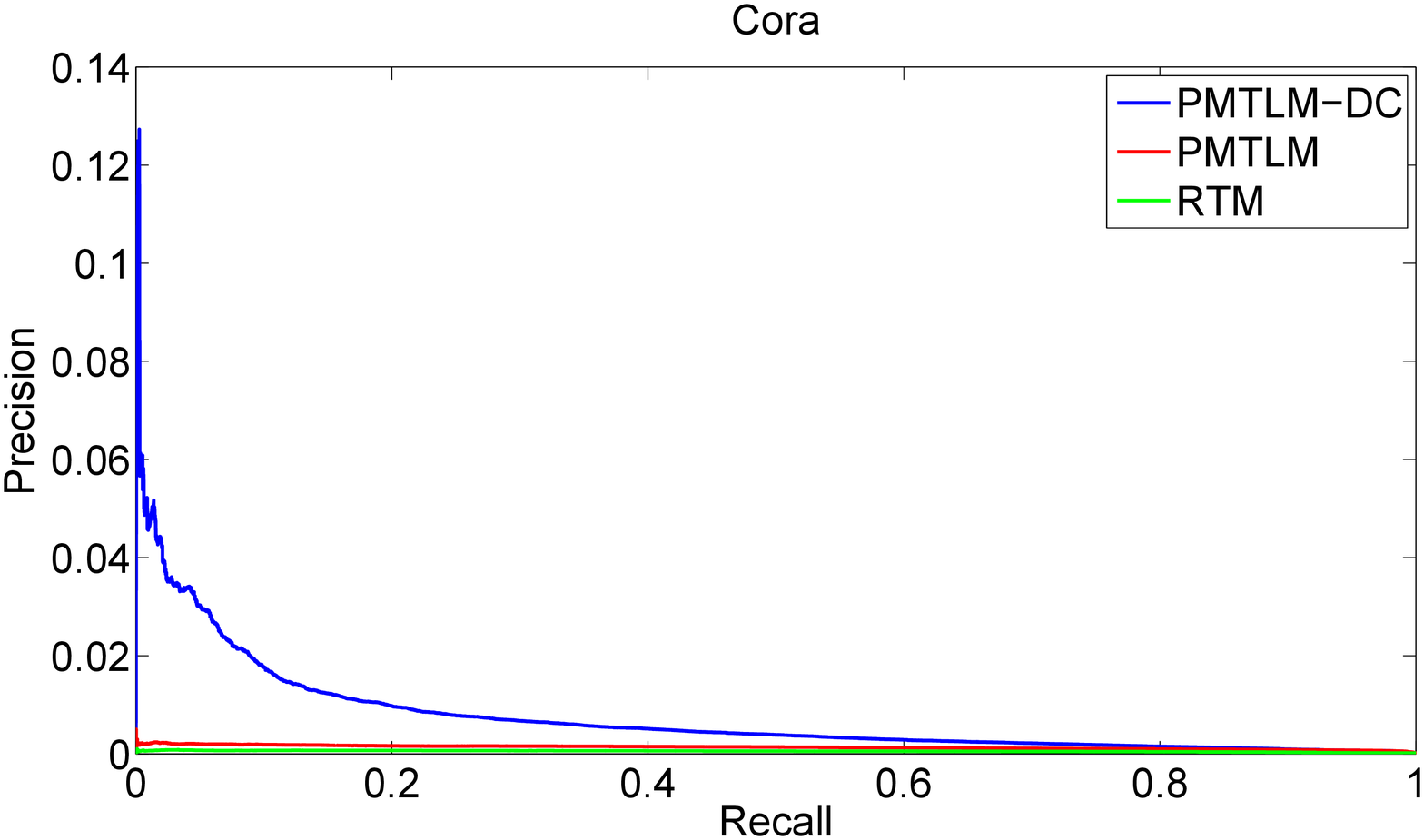,width=0.32 \textwidth, height=3.3cm}
\epsfig{file=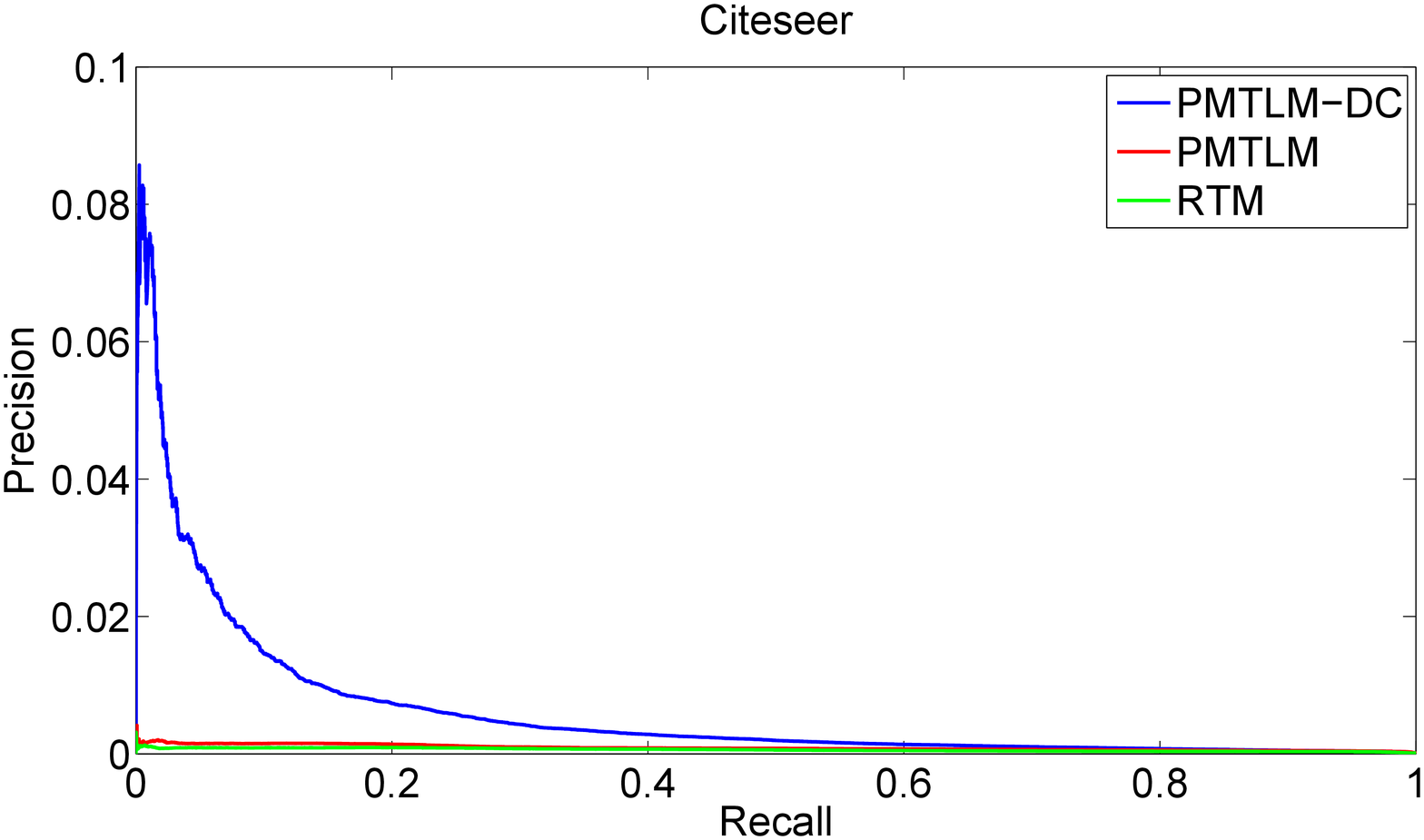,width=0.32 \textwidth, height=3.3cm}
\epsfig{file=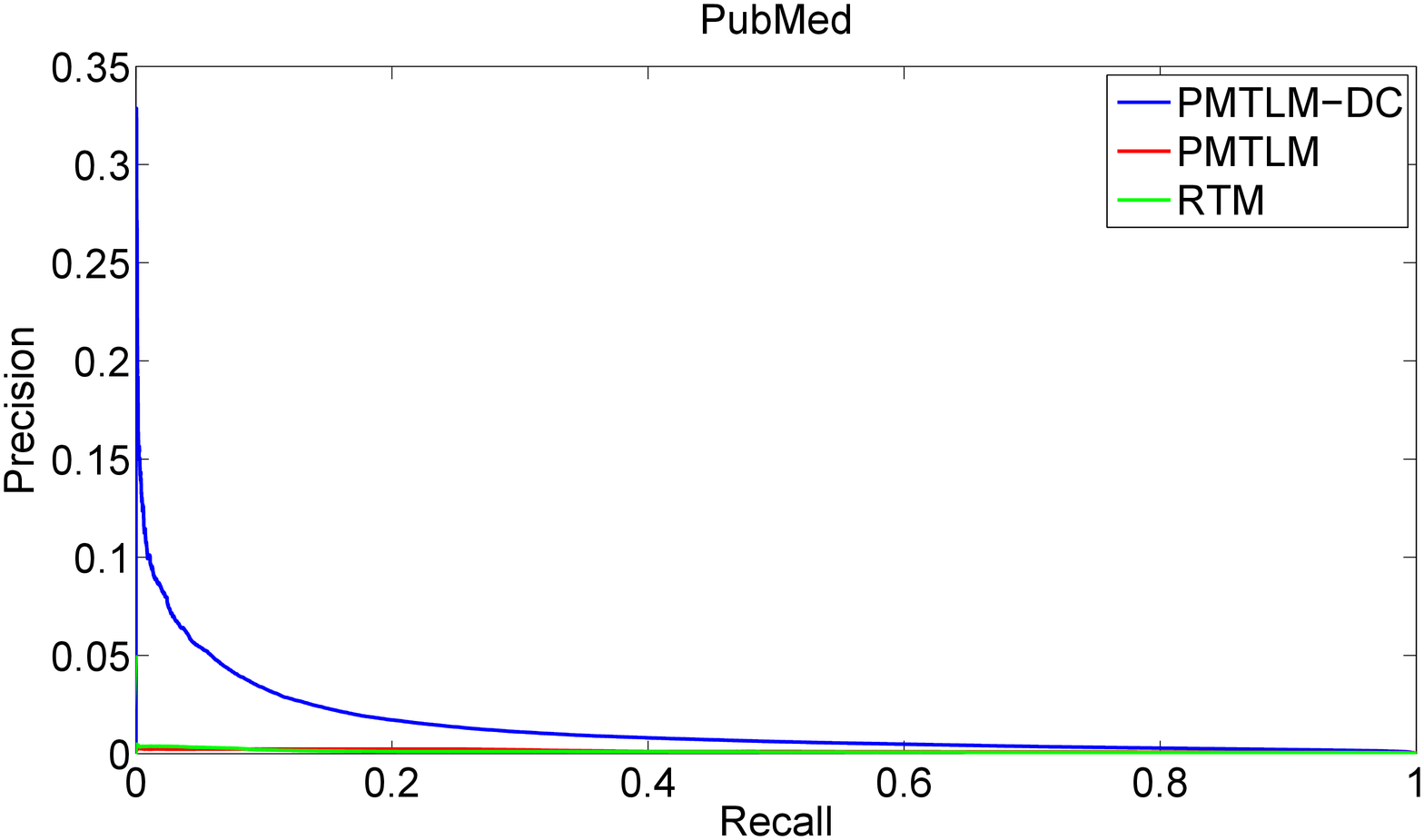,width=0.32 \textwidth, height=3.3cm}
\label{fig:PRC}
}
\caption{Performance on the link prediction task.  For all three data sets and all the $\alpha$ values, the \mbox{PMTLM-DC} model achieves higher accuracy than the \mbox{PMTLM} model.
In contrast to~\figref{fig:weight_NMI_cora_citeseer}, for this task the optimal value of $\alpha$ is relatively small, showing that the content is less important, and the topology is more important, for link prediction than for document classification.  The green line in~\figref{fig:AUC} indicates the highest AUC achieved by the \mbox{RTM} model, maximized over the tunable parameter $\rho$.  Our models outperform \mbox{RTM} on all three data sets.  In addition, the degree-corrected model (\textsc{PMTLM-DC}) does significantly better than the uncorrected version (\textsc{PMTLM}).
\label{fig:AUC-ROC-PRC}}
\end{figure*}

\figref{fig:AUC} gives the \mbox{AUC} values for \MTLM\ and \MTLMDC\ as a function of the relative weight $\alpha$ of content vs.\ links. The green horizontal line in each of those subplots represent the highest \mbox{AUC} value achieved by the \RTM\ model for each data set, using the best value of $\rho$ among those specified in~\secref{sec:doc_classification}.   Interestingly, for Cora and Citeseer the optimal value of $\alpha$ is smaller than in~\figref{fig:weight_NMI_cora_citeseer}, showing that content is less important for link prediction than for document classification.  We also plot the \emph{receiver operating characteristic} (\mbox{ROC}) curves and precision-recall curves that achieve the highest \mbox{AUC} values in~\figref{fig:ROC} and~\figref{fig:PRC} respectively.

We see that, for all three data sets, our models outperform \RTM, and that the degree-corrected model \MTLMDC\ is significantly more accurate than the uncorrected one.

\section{Conclusions}
\label{sec:conclusion}
We have introduced a new generative model for document networks.  It is a marriage between Probabilistic Latent Semantic Analysis~\cite{Hofmann1999} and the Ball-Karrer-Newman mixed membership block model~\cite{Ball2011}.  Because of its mathematical simplicity, its parameters can be inferred with a particularly simple and scalable EM algorithm.  Our experiments on both document classification and link prediction show that it achieves high accuracy and efficiency for a variety of data sets, outperforming a number of other methods.  In future work, we plan to test its performance for other tasks including supervised and semisupervised learning, active learning, and content prediction, i.e., predicting the presence or absence of words in a document based on its links to other documents and/or a subset of its text.


\section{Acknowledgments}

We are grateful to Brian Ball, Brian Karrer, Mark Newman and David M. Blei for helpful conversations.  Y.Z., X.Y., and C.M. are supported by AFOSR and DARPA under grant FA9550-12-1-0432.

%
\bibliographystyle{abbrv}
\bibliography{draft}  
%
%
\appendix
\section{Update Equations for \mbox{PMTLM}}
\label{sec:appedix_a}
In this appendix, we derive the update equations~\eqref{eq:mle_eta}--\eqref{eq:mle_theta} for the parameters $\eta$, $\beta$, and $\theta$, giving the M step of our algorithm.

Recall that the likelihood is given by~\eqref{eq:weight_norm_llh} and~\eqref{eq:llh_jensen}.
For identifiability, we impose the normalization constraints
\begin{align}
\forall z: \sum_w \beta_{zw}  & = 1 \label{eq:constraint-beta} \\
\forall d: \sum_z \theta_{dz}  & = 1 \label{eq:constraint-theta}
\end{align}
For each topic $z$, taking the derivative of the likelihood with respect to $\eta_z$ gives
\begin{align}
0 = \frac{1}{1-\alpha} \frac{\partial \mathcal{L}}{\partial \eta_z}
 &= \frac{1}{\eta_z} \sum_{dd'} A_{dd'} q_{dd'}(z) - \sum_{dd'} \theta_{dz} \theta_{d'z} \, .
\end{align}
Thus
\begin{align}
 \eta_z
 = \frac{\sum_{dd'} A_{dd'} q_{dd'}(z)}{\sum_{dd'} \theta_{dz} \theta_{d'z}}
 = \frac{\sum_{dd'} A_{dd'} q_{dd'}(z)}{\left(\sum_d \theta_{dz} \right)^2} \, .
\end{align}
Plugging this in to~\eqref{eq:llh_jensen} makes the last term a constant, $-1/2 \sum_{dd'} A_{dd'} = -M$.  Thus we can ignore this term when estimating $\theta_{dz}$.

Similarly, for each topic $z$ and each word $w$, taking the derivative with respect to $\beta_{zw}$ gives
\begin{align}
\nu_z = \frac{1}{\alpha} \frac{\partial \mathcal{L}}{\partial \beta_{zw}}
 &= \frac{1}{\beta_{zw}} \,\sum_d \frac{1}{L_d} \,C_{dw} h_{dw}(z) \, ,
\end{align}
where $\nu_z$ is the Lagrange multiplier for~\eqref{eq:constraint-beta}.  Normalizing $\beta_z$ determines $\nu_z$, and gives
\begin{align}
  \label{eq:update_beta_vanilla}
 \beta_{zw}
 = \frac{\sum_d (1/L_d) C_{dw} h_{dw}(z)}{\sum_d (1/L_d) \sum_{w'} C_{dw'} h_{dw'}(z)} \, .
\end{align}

Finally, for each document $d$ and each topic $z$, taking the derivative with respect to $\theta_{dz}$ gives
\begin{align}
\lambda_d = \frac{\partial \mathcal{L}}{\partial \theta_{dz}}
 &= \frac{\alpha}{L_d \theta_{dz}}\sum_w C_{dw} h_{dw}(z)+\frac{1-\alpha}{\theta_{dz}} \sum_{d'} A_{dd'} q_{dd'}(z)
  \, ,
\end{align}
where $\lambda_d$ is the Lagrange multiplier for~\eqref{eq:constraint-theta}.  Normalizing $\theta_d$ determines $\lambda_d$ and gives
\begin{align}
 \label{eq:theta-vanilla}
 \theta_{dz}=\frac{(\alpha/L_d)\sum_{w}C_{dw}h_{dw}(z)+(1-\alpha)\sum_{d'}A_{dd'}q_{dd'}(z)}{\alpha+(1-\alpha)\kappa_d}
  \, .
\end{align}




\section{Update Equations for the Degree-Corrected model}
\label{sec:appedix_b}

Recall that in the degree-corrected model \MTLMDC, the number of links between each pair of documents $d, d'$ is Poisson-distributed with mean
\begin{equation}
S_d S_{d'} \sum_z \eta_z \theta_{dz} \theta_{d'z} \, .
\end{equation}
To make the model identifiable, in addition to~\eqref{eq:constraint-beta} and~\eqref{eq:constraint-theta}, we impose the following constraint on the degree-correction parameters,
\begin{align}
\forall z: \sum_d S_d \theta_{dz} & = 1 \label{eq:constraint-S} \, .
\end{align}
With this constraint, we have
\begin{align}
\mathcal{L}
&= \alpha \sum_{d}\frac{1}{L_d}\sum_{wz} C_{dw} h_{dw}(z) \log \frac{\theta_{dz} \beta_{zw}}{h_{dw}(z)}\nonumber \\
&+ (1-\alpha)\sum_d \kappa_d \log S_d\nonumber\\
&+ \frac{1-\alpha}{2} \sum_{dd'z} \left( A_{dd'} q_{dd'}(z) \log \frac{\eta_z \theta_{dz} \theta_{d'z}}{q_{dd'}(z)} - S_d S_{d'} \eta_z \theta_{dz} \theta_{d'z} \right)\,.
\end{align}
The update equation~\eqref{eq:update_beta_vanilla} for $\beta$ remains the same, since the degree-correction only affects the part of the model that generates the links, not the words.
We now derive the update equations for $\eta$, $S$, and $\theta$.



For each topic $z$, taking the derivative of the likelihood with respect to $\eta_z$ gives
\begin{align}
\label{eq:diff-eta}
0 = \frac{2}{1-\alpha} \frac{\partial L}{\partial \eta_z}
&= \frac{1}{\eta_z} \sum_{dd'} A_{dd'} q_{dd'}(z) - \sum_{dd'} S_d S_{d'} \theta_{dz} \theta_{d'z} \nonumber \\
&= \frac{1}{\eta_z} \sum_{dd'} A_{dd'} q_{dd'}(z) - 1 \, ,
\end{align}
where we used~\eqref{eq:constraint-S}.  Thus
\begin{equation}
\label{eq:update-eta}
\eta_z = \sum_{dd'} A_{dd'} q_{dd'}(z) \, ,
\end{equation}
so $\eta_z$ is simply the expected number of links caused by topic $z$.  In particular,
\begin{equation}
\label{eq:total-eta}
\sum_z \eta_z = \sum_{dd'} A_{dd'} = \sum_d \kappa_d = 2M \, .
\end{equation}
For $S_d$, we have
\begin{align}
\frac{1}{1-\alpha} \frac{\partial L}{\partial S_d}
&= \frac{\kappa_d}{S_d} - \sum_{d'z} S_{d'} \eta_z \theta_{dz} \theta_{d'z}
\nonumber \\
&= \frac{\kappa_d}{S_d} - \sum_{z} \eta_z \theta_{dz} = \sum_z \xi_z \theta_{dz} \, ,
\label{eq:diff-S}
\end{align}
where $\xi_z$ is the Lagrange multiplier for~\eqref{eq:constraint-S}.  Thus
\begin{equation}
\label{eq:update-S}
S_d = \frac{\kappa_d}{\sum_z (\eta_z + \xi_z) \theta_{dz}} \, .
\end{equation}
We will determine $\xi_z$ below.  However, note that multiplying both sides of~\eqref{eq:diff-S} by $S_d$, summing over $d$, and applying~\eqref{eq:constraint-S} and~\eqref{eq:total-eta} gives
\begin{equation}
\sum_z \xi_z
= 0 \, .
\end{equation}
Most importantly, for $\theta$ we have
\begin{align}
\frac{\partial L}{\partial \theta_{dz}}
&= \frac{1}{\theta_{dz}}
\left(
\frac{\alpha}{L_d} \sum_w C_{dw} h_{dw}(z)
+ (1-\alpha) \sum_{d'} A_{dd'} q_{dd'}(z)
\right)
\nonumber \\
&- (1-\alpha) \sum_{d'} S_d S_{d'} \eta_z \theta_{d'z} \nonumber \\
&= \frac{1}{\theta_{dz}}
\left(
\frac{\alpha}{L_d} \sum_w C_{dw} h_{dw}(z)
+ (1-\alpha) \sum_{d'} A_{dd'} q_{dd'}(z)
\right)\nonumber\\
&- (1-\alpha) S_d \eta_z \nonumber \\
&= \lambda_d + (1-\alpha) S_d \xi_z \, ,
\label{eq:diff-theta}
\end{align}
where $\lambda_d$
is the Lagrange multiplier for~\eqref{eq:constraint-theta}, and where we applied~\eqref{eq:constraint-S} in the second equality.
Multiplying both sides of~\eqref{eq:diff-theta} by $\theta_{dz}$, summing over $z$, and applying~\eqref{eq:update-S} gives
\begin{equation}
\label{eq:lambda}
\lambda_d = \alpha \, .
\end{equation}
Summing over $d$ and applying~\eqref{eq:constraint-S}, \eqref{eq:update-eta}, and~\eqref{eq:lambda} gives
\begin{align}
\label{eq:xi}
 \frac{1-\alpha}{\alpha} \,\xi_z
 &= \sum_{d} \frac{1}{L_d} \,\sum_{w} C_{dw} h_{dw}(z) - \sum_d \theta_{dz}
 \nonumber \\
 &= \sum_{d}\frac{1}{L_d} \,\sum_{w} C_{dw} \left( h_{dw}(z) - \theta_{dz} \right)
 \, .
\end{align}
Thus $\xi_z$ measures how the inferred topic distributions of the words $h_{dw}(z)$ differ from the topic mixtures $\theta_{dz}$.

Finally, \eqref{eq:diff-theta} and~\eqref{eq:lambda} give
\begin{align}
\label{eq:update-theta}
\theta_{dz}
&= \frac{
(\alpha/L_d) \sum_w C_{dw} h_{dw}(z)
+ (1-\alpha) \sum_{d'} A_{dd'} q_{dd'}(z)
}
{\alpha + (1-\alpha) (\eta_z + \xi_z) S_d} \, ,
\end{align}
where $\eta_z$ and $\xi_z$ are given by \eqref{eq:update-eta} and~\eqref{eq:xi}.



\section{Update Equations with Dirichlet Prior}
\label{sec:appedix_c}

If we impose a Dirichlet prior on $\theta$, with parameters $\{\gamma_z\}$ for each topic $z$, this gives an additional term $\sum_{dz}(\gamma_z-1)\log\theta_{dz}$ in the log-likelihood of both the \MTLM\ and \MTLMDC\ models. This is equivalent to introducing pseudocounts $t_z = \gamma_z-1$ for each $z$, which we can think of as additional words or links that we know are due to topic $z$.  Our original models, without this term, correspond to the uniform prior with $\gamma_z = 1$ and $t_z = 0$.  However, as long as $\gamma_z \ge 1$ so that the pseudocounts are nonnegative, we can infer the parameters of our model in the same way with no loss of efficiency.

In the \MTLM\ model,~\eqref{eq:theta-vanilla} becomes
\begin{align}
 \theta_{dz}
 = \frac{t_z + (\alpha/L_d)\sum_{w}C_{dw}h_{dw}(z)+(1-\alpha)\sum_{d'}A_{dd'}q_{dd'}(z)}
 {\sum_z t_z + \alpha + (1-\alpha)\kappa_d}
  \, .
\end{align}
In the degree-corrected model \MTLMDC,~\eqref{eq:lambda} and~\eqref{eq:xi} become
\begin{equation}
\label{eq:lambda-prior}
\lambda_d =  \alpha + \sum_z t_z
\end{equation}
and
\begin{align}
\label{eq:xi-prior}
\frac{1-\alpha}{\alpha} \,\xi_z
&= \sum_d \frac{1}{L_d} \,\sum_w C_{dw} \left(h_{dw}(z)-\theta_{dz} \right) \nonumber \\
&+ \frac{1}{\alpha} \sum_d \left(
t_z - \theta_{dz} \sum_{z'} t_{z'} \right)
\, .
\end{align}
Note that $\xi_z$ has two contributions.  One measures, as before, how the inferred topic distributions of the words $h_{dw}(z)$ differ from the topic mixtures $\theta_{dz}$, and the other measures how the fraction $t_z / \sum_{z'} t_{z'}$ of pseudocounts for topic $z$ differs from $\theta_{dz}$.

Finally,~\eqref{eq:update-theta} becomes
\begin{align}
\label{eq:update-theta-prior}
\theta_{dz}
&= \frac{t_z
+ (\alpha/L_d) \sum_w C_{dw} h_{dw}(z)
+ (1-\alpha) \sum_{d'} A_{dd'} q_{dd'}(z)
}
{\alpha + (1-\alpha) (\eta_z + \xi_z) S_d + \sum_{z'} t_{z'}} \, ,
\end{align}
where $\eta_z$ and $\xi_z$ are given by~\eqref{eq:update-eta} and~\eqref{eq:xi-prior}.

\end{document}